\documentclass[10pt,journal,compsoc]{IEEEtran}

\usepackage{graphicx}
\usepackage{booktabs}

\usepackage{tikz}
\usepackage{comment}
\usepackage{amsmath,amssymb} 
\usepackage{color}

\usepackage{multirow}
\usepackage{url}
\usepackage{ulem}
\usepackage{algorithm} 
\usepackage{algpseudocode}

\usepackage{hyperref}
\usepackage{amsmath}
\algblock{Input}{EndInput}
\algnotext{EndInput}
\algblock{Output}{EndOutput}
\algnotext{EndOutput}
\newcommand{\Desc}[2]{\State \makebox[2em][l]{#1}#2}
\usepackage[inline]{enumitem}

\usepackage{balance} 

\usepackage{pifont}
\newcommand{\cmark}{\ding{51}}%
\newcommand{\xmark}{\ding{55}}%
\newcommand{\red}[1]{\textcolor{black}{#1}}

\usepackage{multicol}
\usepackage{subcaption}
\newcommand{\customtitle}{
    \begin{center}
        {\LARGE \bfseries Supplementary Material for \\
Every Angle Is Worth A Second Glance: Mining Kinematic Skeletal Structures from Multi-view Joint Cloud} \\[1em]
        {\large Junkun~Jiang, Jie~Chen,~\IEEEmembership{Member,~IEEE}, Ho~Yin~Au,~Mingyuan~Chen,~Wei~Xue,~\IEEEmembership{Member,~IEEE}, and~Yike~Guo,~\IEEEmembership{Senior Member,~IEEE}} \\[1em]
    \end{center}
}

\usepackage[accsupp]{axessibility}  

\ifCLASSOPTIONcompsoc
  \usepackage[nocompress]{cite}
\else
  \usepackage{cite}
\fi

\begin{document}
%
\title{Every Angle Is Worth A Second Glance: \\Mining Kinematic Skeletal Structures \\from Multi-view Joint Cloud}

%
%

\author{Junkun~Jiang, Jie~Chen,~\IEEEmembership{Senior Member,~IEEE}, Ho~Yin~Au,~Mingyuan~Chen,~Wei~Xue,~\IEEEmembership{Member,~IEEE}, and~Yike~Guo,~\IEEEmembership{Fellow,~IEEE}
\IEEEcompsocitemizethanks{\IEEEcompsocthanksitem J. Jiang, J. Chen (corresponding author), H. Y. Au, and M. Chen are with the Department of Computer Science, Hong Kong Baptist University, Hong Kong SAR, China.
(e-mails: \{csjkjiang, chenjie, cshyau\}@comp.hkbu.edu.hk, csmychen@hkbu.edu.hk).
\IEEEcompsocthanksitem W. Xue is with the Division of Emerging Interdisciplinary Areas, the Hong Kong University of Science and Technology, Hong Kong SAR, China. (e-mail: weixue@ust.hk).
\IEEEcompsocthanksitem Y. Guo is with the Department of Computer Science and Engineering, the Hong Kong University of Science and Technology, Hong Kong SAR, China. (e-mail: yikeguo@ust.hk).
\IEEEcompsocthanksitem This work was supported by the Research Grants Council of Hong Kong (Project No. T45-205/21-N), and by the Hong Kong Baptist University (Project No. RC-OFSGT2/20-21).
}
}

%
%

\markboth{}%
{Shell \MakeLowercase{\textit{et al.}}: Bare Demo of IEEEtran.cls for Computer Society Journals}
%



\IEEEtitleabstractindextext{%
\begin{abstract}
Multi-person motion capture over sparse angular observations is a challenging problem under interference from both self- and mutual-occlusions. Existing works produce accurate 2D joint detection, however, when these are triangulated and lifted into 3D, available solutions all struggle in selecting the most accurate candidates and associating them to the correct joint type and target identity. 
As such, in order to fully utilize all accurate 2D joint location information, we propose to independently triangulate between all same-typed 2D joints from all camera views regardless of their target ID, forming the Joint Cloud. Joint Cloud consist of both valid joints lifted from the same joint type and target ID, as well as falsely constructed ones that are from different 2D sources. These redundant and inaccurate candidates are processed over the proposed Joint Cloud Selection and Aggregation Transformer (JCSAT) involving three cascaded encoders which deeply explore the trajectile, skeletal structural, and view-dependent correlations among all 3D point candidates in the cross-embedding space. An Optimal Token Attention Path (OTAP) module is proposed which subsequently selects and aggregates informative features from these redundant observations for the final prediction of human motion. To demonstrate the effectiveness of JCSAT, we build and publish a new multi-person motion capture dataset BUMocap-X with complex interactions and severe occlusions. Comprehensive experiments over the newly presented as well as benchmark datasets validate the effectiveness of the proposed framework, which outperforms all existing state-of-the-art methods, especially under challenging occlusion scenarios.
\end{abstract}

\begin{IEEEkeywords}
3D human pose estimation, motion capture, transformer, optimal token attention path
\end{IEEEkeywords}}

\maketitle

\IEEEdisplaynontitleabstractindextext

%
\IEEEpeerreviewmaketitle

\section{Introduction}\label{sec:introduction}
\IEEEPARstart{W}{ith} the fast development of Virtual Reality (VR) and Augmented Reality (AR) techniques, accessible VR/AR devices~\cite{vive,hololens,oculus} allow people to control their own avatars immersively by capturing body motion from a headset with two handheld controllers. On the other hand, commercial Motion Capture systems~\cite{optitrack,vicon,xsens} provide high-precision solutions in animation and film-making areas. The requirement of wearing pieces of equipment leads to inconvenient experiences. To ease those issues, researchers propose device-free motion-capturing frameworks. 
One of the most efficient and popular approaches is Vision-based Human Pose Estimation (HPE), showing remarkable progress in 2D multi-person pose detections~\cite{cao2019openpose,wu2019detectron2,sun2019deep,fang2022alphapose} and 3D single-person pose estimations~\cite{pavllo20193d,kolotouros2019learning,jiang2024explore}. However, motion capture usually involves multiple targets in one scene, and thus introduces occlusions and prevents accurate movement estimation. Multi-view observations are then introduced to provide additional information based on the assumption that joints occluded from one view are visible from another view.  \red{Current studies} \cite{zhang20204d,dong2021fast,tu2020voxelpose,kolotouros2019learning,li2022exploiting,zhou2022quickpose} show that multi-view multi-person 3D HPE can be formulated as a multi-stage framework, i.e., use a 2D pose detector to estimate 2D poses from calibrated observations, and then apply triangulation algorithm~\cite{hartley2003multiple} to reconstruct 3D motion.

\begin{figure}[t]
\centering
\setlength{\abovecaptionskip}{0mm}
\includegraphics[width=0.98\linewidth]{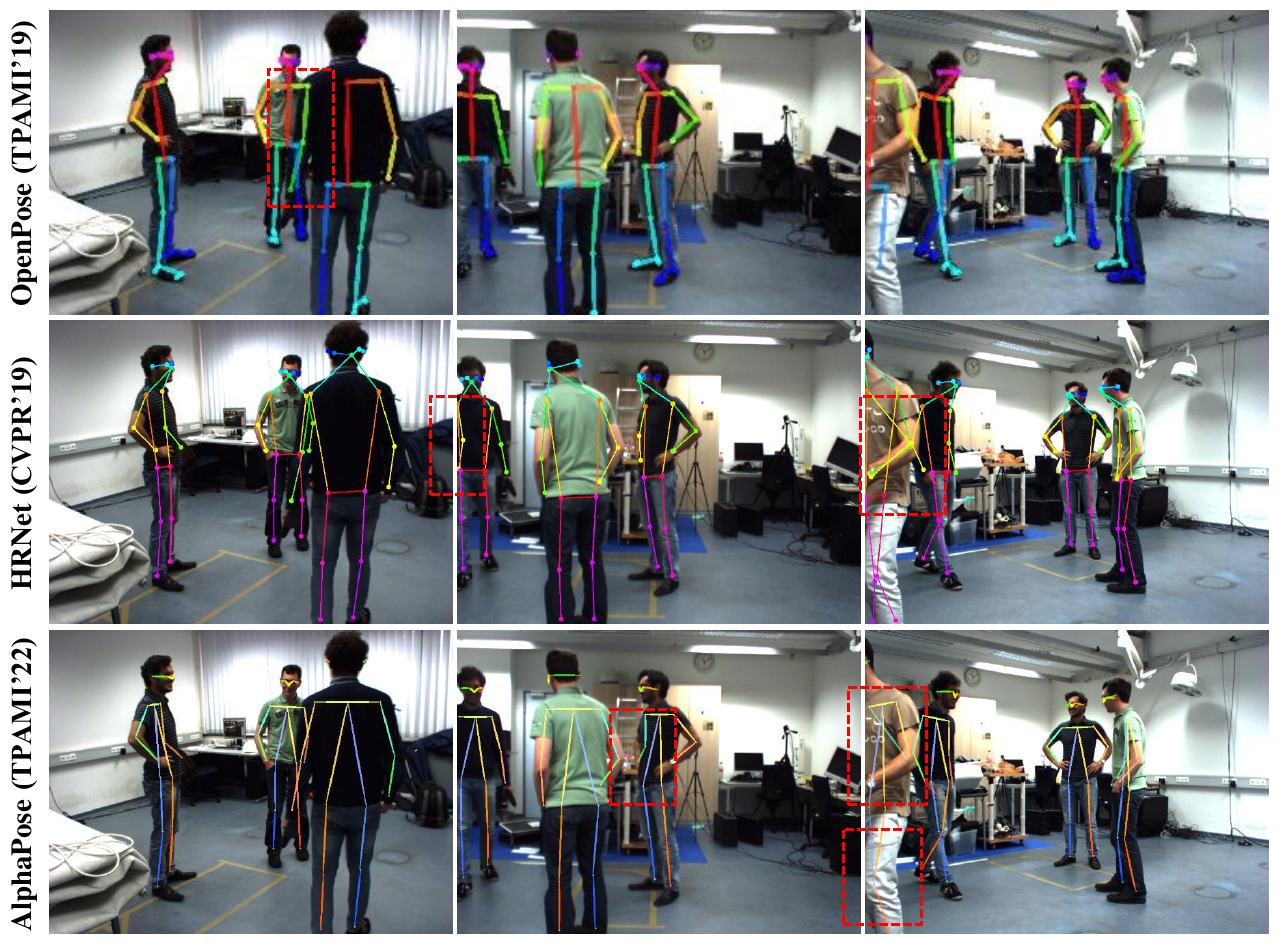}
\caption{Demonstration of 2D detections from state-of-the-art 2D pose detectors~\protect\cite{cao2019openpose,fang2022alphapose,sun2019deep} at the same timestamp on Shelf dataset~\protect\cite{belagiannis2014multiple}. Although both produce the most accurate results, there are still some defects notated by red dotted boxes regarding the limbs: 1) assigned to incorrect target identity, 2) assigned to incorrect joint type.}
\vspace{-5mm}
\label{fig_2d_detection}
\end{figure}

The main challenge of a multi-stage framework is the association of 2D detections across views to the target identity and joint type. Due to ambiguities caused by occlusions, the inaccuracy in 2D detections may lead to bad associations. Here we categorise occlusions into self-occlusion which is caused by one's own body part, and mutual-occlusion which is caused by others. As shown in Fig.~\ref{fig_2d_detection}, 2D detectors have inherent limitations on occlusions: 1) self-occlusion happened in Row 2, Col 3, makes the left knee assigned to the wrong joint type (right knee), 2) mutual-occlusion happened in Row 1, Col 1, makes the left arm assigned to the wrong target ID (other people). Incorrect association error accumulates throughout the whole lifetime of the multi-stage framework, producing twisted motion results.
To solve those issues, existing approaches~\cite{zhang20204d,dong2021fast,tu2020voxelpose,reddy2021tessetrack,wu2021graph} select the most accurate 2D candidates (discarding others) and then \red{associate these with the target ID and joint type}. 
\red{However, these} may lead to sub-optimal results \red{(incorrect target ID and joint type)} and cannot truly solve the error-accumulation issue.
\red{In our previous study~\cite{jiang2022dual}, we propose to utilize the masking learning mechanism~\cite{he2022masked} to explore the correlation among partial motion. It enables us to circumvent the above sub-optimal issue by filtering those low-confidence 3D joints and reconstructing the complete motion with high accuracy performance and also enables our model to handle the missing. Nevertheless, as analyzed, 2D detections may have the wrong ID/joint type but accurate positions, which we cannot overlook.}

\begin{figure*}
\centering
\setlength{\abovecaptionskip}{-1mm}
\includegraphics[width=0.98\textwidth]{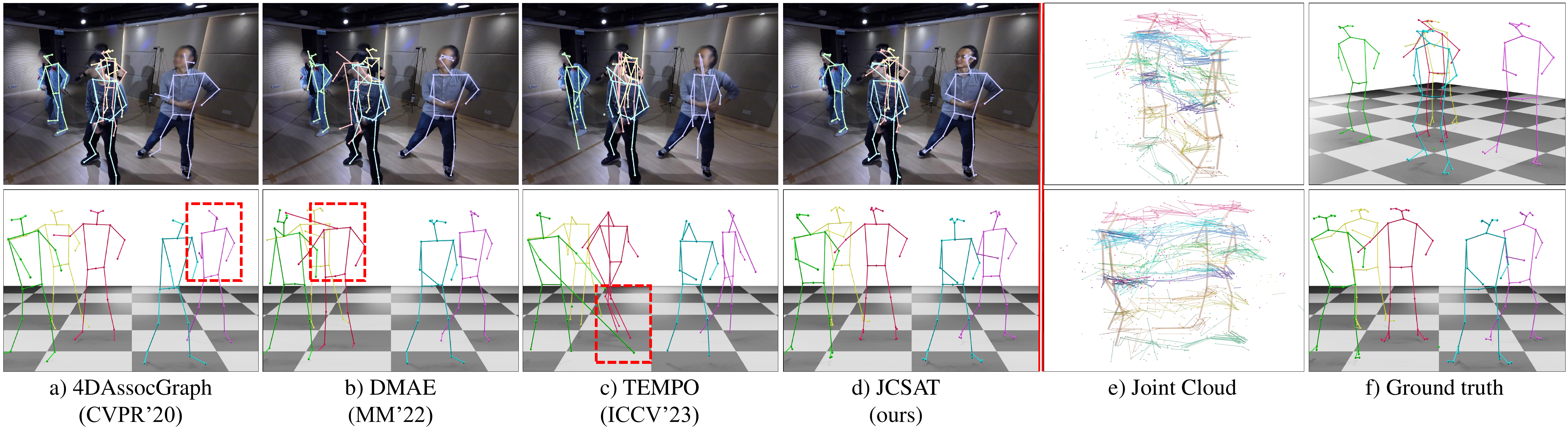}
\caption{Qualitative comparison with the most recent multi-view multi-person motion capture frameworks on our \red{newly collected} dataset \textbf{BUMocap-X}. We demonstrate the projection and rendered results as well as the manual-labelled ground truth (from left to right: 4DAssocGraph~\cite{zhang20204d}, DMAE~\cite{jiang2022dual}, TEMPO~\cite{choudhury2023tempo}, ours, the Joint Cloud, and the ground truth). We provide the rendered Joint Cloud of Actor-3 in ten frames. Same-typed joints are connected to indicate the trajectory. The proposed model reasons on trajectile, structural skeletal and angular information from the Joint Cloud and \red{produces} better predictions than existing frameworks, especially under occlusion scenarios.}
\label{fig:teaser}
\end{figure*}

\red{To this end, our approach aims to leverage all available 2D detections by operating under the assumption that each detected 2D joint possesses an accurate position, but with a lower certainty regarding its target ID and joint type}. \red{Following this}, we introduce \red{the concept of} \textbf{\textit{Joint Cloud (JC)}}, which consists of all 3D triangulated candidates from 2D detections within the same joint type between every two views, as shown in Fig.~\ref{fig:teaser}. Therefore JC contains both potentially valid joints lifted from the same target ID and joint type, as well as irregular ones that may come from different sources. 
We further propose a Joint Cloud Selection and Aggregation Transformer (JCSAT) based on three cascaded encoders \red{that} deeply explore the trajectile, skeleton-structural, and view-dependent correlations among all 3D candidates in the cross-embedding space. Besides, an Optimal Token Attention Path (OTAP) module is proposed to select and aggregate informative features from redundant candidates. 
%
In addition, following the masked language modelling task in BERT~\cite{devlin2018bert}, we train our model based on a masked learning scheme to increase \red{its} robustness. 
%
Experiments on public datasets demonstrate the efficiency of our proposed framework against state-of-the-art methods. We also validate our method on a \red{newly collected} dataset\red{, BUMocap-X which features} complex interactions and heavy occlusions.
\red{BUMocap-X is an expanded version of BUMocap~\cite{jiang2022dual}, approximately 0.3 times larger and includes two additional actors' motions. It captures a 5-actor group dance with complex interaction.}

The main contributions are listed as follows: 

\begin{itemize}
\item We introduce the concept, formation, and exploitation methodology of the Joint Cloud (JC). JC is designed to fully utilize all accurate 2D joint locations via independent triangulation between all same-typed 2D joints from all camera views regardless of their target ID. This data arrangement improves information utilization from all observation angles.
\item We propose the Joint Cloud Selection and Aggregation Transformer (JCSAT) framework which involves three cascaded encoders that deeply explore the trajectile, skeleton-structural, and view-dependent correlations among all 3D candidates in the cross-embedding space.
\item Different from existing Transformer frameworks that rely on full self-attention mechanisms to associate and aggregate features, we propose the Optimal Token Attention Path (OTAP) module which selects the most informative feature from noisy and redundant options for more robust and accurate prediction of human motion.
\item \red{We build and publish the BUMocap-X dataset, a comprehensively and precisely annotated multi-person motion capture dataset featuring complex interactions and significant occlusions. Utilizing this dataset, we showcase the robustness of our proposed framework in effectively handling challenging occlusion scenarios}.
\end{itemize}

\section{Related Works}

\noindent\textbf{Multi-view multi-person 3D HPE.} 
\red{These frameworks are normally built upon two stages.}
We divide them into two categories: optimization-based and learning-based. 
Optimization-based methods mainly focus on the second stage.
Pioneers in this field, such as \cite{belagiannis20143d,belagiannis2014multiple} propose a 3D pictorial structure (3DPS) model for searching \red{corresponding} body joints from a common state space shared by all bodies. 
Zhang et al.~\cite{zhang20204d} associate 2D joints across views by optimizing the proposed geometric energy function, which includes the epipolar distance, PAF score~\cite{cao2019openpose} and tracking distance. 
Dong et al.~ \cite{dong2021fast} propose two affinity matrices for associating target bodies. These matrices measure the distance between 2D input pairs in terms of re-id visual features and geometric constraints. They adopt the 3DPS model to reconstruct the final motion. 
Inspired by \cite{zhang20204d,dong2021fast}, Zhou et al.~\cite{zhou2022quickpose} associate joint at part- and body-level. They build a skeletal graph, which includes all possible joint connections, and adopt a clustering algorithm to optimize the best connection.
Learning-based methods~\cite{tu2020voxelpose,reddy2021tessetrack} propose to use 3D convolutional neural networks (CNNs) to locate the most likely 3D joint from a voxel-based 3D heatmap. Those techniques allow for an end-to-end \red{manner} where losses can be back propagated to the first stage. However, those voxel-based regressions have a significant computing cost due to their greedy space discretization. \red{Following the voxel representation, Ye et al.~\cite{ye2022faster} accelerate the inference speed by utilizing 2D CNNs. Choudhury et al.~\cite{choudhury2023tempo} apply a Kalman filter and Spatial Gated Recurrent Units (GRUs) for pose tracking and forecasting. SelfPose3D~\cite{srivastav2024selfpose3d} suggests that combining synthetic root data with augmented 2D pseudo-pose data improves model accuracy. Despite their high accuracy, these methods lack generalizability and are unable to cope with changes in camera parameters. On the other hand,} Lin and Lee~\cite{lin2021multi} propose a plane-sweep-stereo-based network for regressing each joint's depth in every view. It \red{enables} real-time inference but requires fusing 3D poses from all views into the same coordinate, \red{which may result in duplicate poses}. 
We argue that, regardless of whether the method is optimization- or learning-based, the quality of 2D detections remains the bottleneck of the multi-stage framework.
Under heavy occlusions, 2D detections may have false target ID and joint type (see Fig.~\ref{fig_2d_detection}) in the first stage, resulting in the error-accumulated association in the second stage and ultimately leading to sub-optimal performance. In contrast, we propose the Joint Cloud which consists of all 3D joint candidates regardless of their target ID in order to preserve 2D detections in terms of their precise location. We then propose the JCSAT framework to select the potentially valid ones from redundancy with the trajectile, skeletal and angular cross-embedding. 

\noindent\textbf{Motion prior.} Recently, some research has been done to involve motion prior to 3D HPE tasks. Some studies~\cite{rempe2021humor,huang2022neural} predict the contact of limbs to make the 3D reconstruction physically plausible. Some~\cite{gong2022posetriplet,maeda2022motionaug} use physics simulation to avoid floor penetration. Some other works~\cite{kolotouros2019learning,wandt2022elepose,wandt2021canonpose} propose to estimate each camera's extrinsic parameters based on projection consistency across views. 
\red{Pose2UV~\cite{huang2022pose2uv} focuses on human mesh recovery. It incorporates the human body mesh prior and the 2D pose prior to predict the corresponding SMPL-based~\cite{loper2015smpl} multi-person movements. Notably, it introduces a large-scale multi-view multi-person motion dataset \textbf{3DMPB} with automatically annotated human target segmentation maps and SMPL data. In contrast, our datasets, BUMocap and BUMocap-X, are annotated manually.}
And we involve a weak prior which comes from the body's kinematic structure. We presume that the bone length should remain consistent throughout a motion sequence, and the structure should maintain a symmetrical form.

\noindent\textbf{Graph-based motion modelling.} Another line of work exemplifies the potential of graph models in human motion modelling. Yan et al.~\cite{yan2018spatial} focus on diverse designs of graph convolution kernels. Various works~\cite{zheng20213d,zhu2021posegtac,li2022exploiting,jiang2022dual} propose implementing Transformer-based networks for 3D HPE tasks. Some~\cite{zheng20213d,zhu2021posegtac,li2022exploiting} treat each input 2D joint as a token~\cite{devlin2018bert} and apply Transformer~\cite{vaswani2017attention} \red{to} learn the projection between 2D and 3D. 
\red{POTR-3D~\cite{park2023towards} proposes to utilize three Transformers to explore multi-person interactions. MTF-Transformer~\cite{shuai2022adaptive} investigates the problem of calibration-free 3D HPE by using Transformers to parallelly predict camera extrinsic and motion.}
\red{Our previous work}~\cite{jiang2022dual} uses a Transformer-based Auto-encoder to complete the 3D motion.
In this work, we further explore the capability of the Transformer in 3D motion modelling. \red{The proposed model receives 3D joint candidates from the Joint Cloud, aggregates and selects them from redundancy, and then regresses these to produce the final motion.}

\begin{figure*}[ht]
\centering
\includegraphics[width=1.0\linewidth]{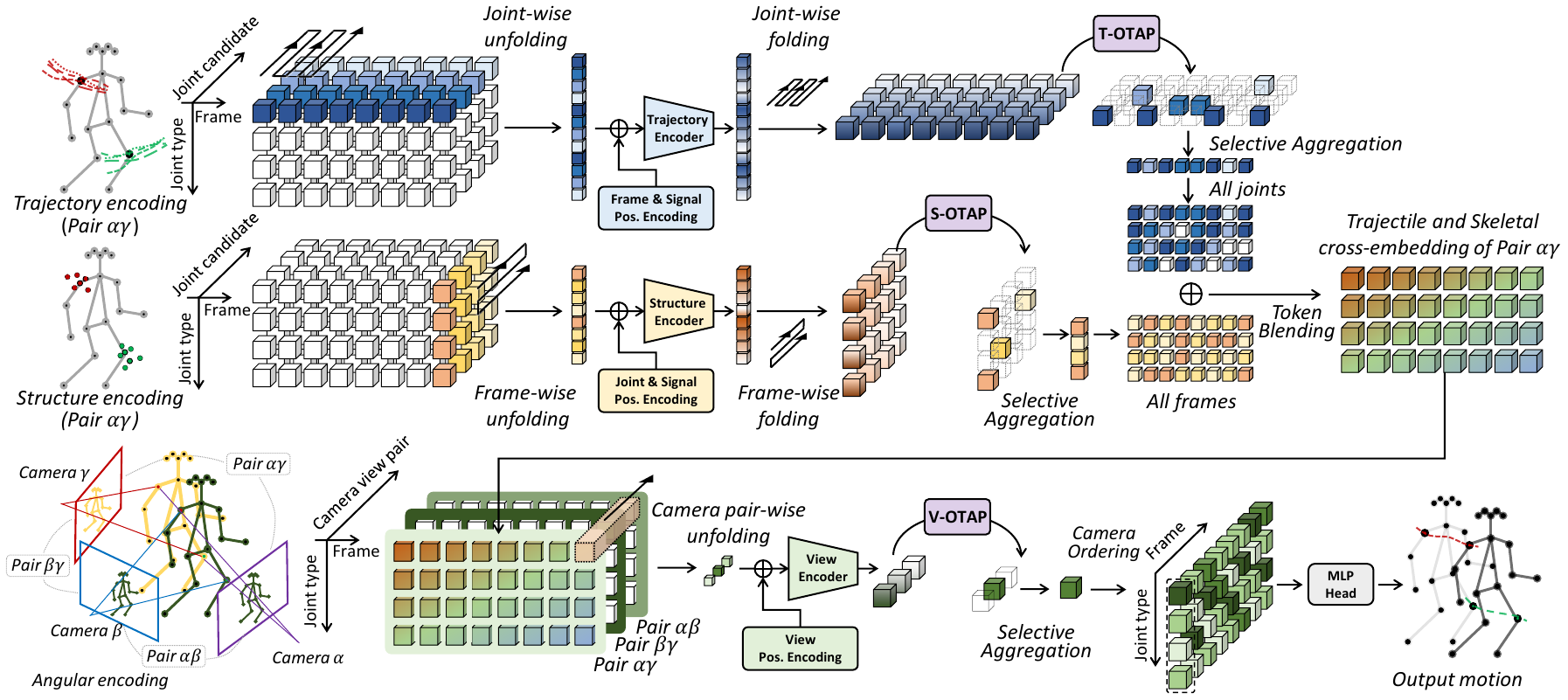}
\caption{The proposed JCSAT pipeline for efficiently extracting kinematic skeletal structures from multi-view Joint Clouds. For every multi-view camera pair (e.g., $\alpha\gamma$), the Trajectory Encoder and Structure Encoder aggregate trajectile and skeletal features from the triangulated Joint Cloud in parallel. 
\red{The T-OTAP and S-OTAP modules select the optimal tokens along the joint candidate axis (joint candidates are constituents of the Joint Cloud, with each joint having multiple candidates triangulated from different camera view pairs) in a differentiable manner.
By traversing through all joints and frames, these candidate feature tokens are then blended element-wise into cross-embedding tokens. 
The cross-embedding tokens are then gathered across the axis of camera view pairs. A View Encoder extracts angular features based on which the subsequent V-OTAP module selects the most representative token for every joint across all frames. Ultimately, an MLP head module decodes the refined trajectory. The three OTAP modules enable comprehensive and efficient optimization of the entire system from end to end.}
}
\label{fig:system}
\end{figure*}


\begin{algorithm}[t]
\caption{Joint Cloud Construction}
\label{alg:Framwork}
\begin{algorithmic}
\Input
\Desc{$ \mathrm{D} $}{, 2D detections}
\Desc{$ \mathrm{C} $}{, Camera matrix}
\EndInput
\Output
\Desc{$\mathcal{J}$}{, Joint Cloud}
\EndOutput
\State 
\Function {Cluster }{$\mathbf{J}, M$}
\State $s \gets \mathbf{J}^{Hip}$
\State $c \gets \mathrm{KMeans}(s, M)$
\For{$m = 1 \to M$ and $i = 1 \to I$} 
\If {$dist(j^{i,m}, c^{m}) < \mathit{th}_i$} \Comment{Thresholding}
\State $\mathbf{J}^m \gets j^{i,m}$
\EndIf
\EndFor
\State \Return{$\mathbf{J}_m$}
\EndFunction
\State 
\Function {Construct Joint Cloud }{$ \mathrm{D}, \mathrm{C} $}
\For{$n = 1 \to N$ and $i = 1 \to I$} 
\For{$(v', v'') \gets \dot{\mathbf{v}} $ and $(m', m'') \gets \dot{\mathbf{m}} $} \Comment{Eq.~\ref{eq.view_comb},~\ref{eq.id_comb}}
\State $j^{i, \dot{\mathbf{m}}}_{n,\dot{\mathbf{v}}} \gets \tau (d_{i,m'}^{n,v'}, d_{i,m''}^{n,v''}, \mathrm{C}_{v'}, \mathrm{C}_{v''})$ \Comment{Triangulate}
\EndFor
\State $\mathbf{J}_{n}^{m} \gets $ \Call{Cluster}{$\mathbf{J}_n, min(M_{n,1}, M_{n,2}, \cdots, M_{n,v})$}
\EndFor
\State \Return{$\mathcal{J}$}
\EndFunction
\end{algorithmic}
\end{algorithm}

\section{Methodology}

Our primary objective is to reconstruct multi-person movements from calibrated multi-view video sequences. 
In contrast to conventional pipelines that discard most of the 2D detections, we try to fully utilize them. To achieve this, we propose the notion of the Joint Cloud in Sec.~\ref{sec:jc}.
Since the Joint Cloud contains \red{potentially valid candidates as well as redundant ones}, we propose a Joint Cloud Selection and Aggregation Transformer (JCSAT)
to distinguish among them in Sec.~\ref{sec:enc}. 
Following each encoder, the redundant features are aggregated by the proposed Optimal Token Attention Path (OTAP) module in Sec.~\ref{sec:jcsa}. 
The aggregated features are last regressed by \red{a multi-layer perception (MLP)} prediction head to 3D motion.
The proposed framework's pipeline is illustrated in Fig.~\ref{fig:system}, and its details are described in the subsequent sub-sections.


\subsection{Joint Cloud Construction}\label{sec:jc}
 

Suppose there are $M$ person targets captured by $V$ calibrated cameras, with each camera recording an $N$ frames of video from different viewing angles.
As the first step, a 2D pose detector (e.g., Openpose~\cite{cao2019openpose}) is applied to estimate 2D poses for each target (person). 
Ideally, the detector correctly identifies $I$ types of human joints (e.g., left shoulder, right elbow and so on) and accurately estimates their 2D locations as $\mathrm{D}_{n,v}=\left\{ d_{n, v}^{i,m} \in \mathbb{R}^2 \right\}$, where subscript $n\in[1, N]$ denotes the frame number and $v\in[1, V]$ denotes the view index, superscript $i\in[1, I]$ denotes the joint type and $m\in[1, M_{n,v}]$ denotes the target identity, respectively. $M_{n,v}$ stands for the total number of observed human targets from the camera view $v$ in the frame $n$. 

\red{As discussed, existing approaches} struggle in associating 2D joints with the correct target ID and joint type consistently across the views. 
Challenges mainly come from ambiguities caused by occlusions \red{resulting in} wrong ID and joint type \red{assignments}.
Unlike conventional pipelines where early evaluations and thresholding operations are applied to exclude ones that are more likely to be wrong and choose only one (or a few) candidates for subsequent regression, we propose to directly triangulate all 2D joints of the same type from all possible pairwise view combinations:
\begin{equation}
\label{eq.view_comb}
\left\{ \dot{\mathbf{v}}=(v', v'')~|~v',v'' \in [1, V]~\text{and}~v'\neq v'' \right\},
\end{equation}
where $\dot{\mathbf{v}}$ represents one of such combinations. Since we do not consider each 2D joint's initial ID information (presumably error-prone under occlusion), the triangulated 3D candidates will have different target ID combinations as well:
\begin{equation}
\label{eq.id_comb}
\left\{\dot{\mathbf{m}}=(m',m'')~|~m' \in [1, M_{n,v'}],~m'' \in [1, M_{n,v''}]\right\}.
\end{equation}

We collect all possible 3D candidates, including those that are extremely noisy and redundant, from each video frame with mixed view and ID combinations, i.e., $\mathbf{J}_n=\left\{j_{n,\dot{\mathbf{v}}}^{i,\dot{\mathbf{m}}} \in \mathbb{R}^3 \right\}$. And by accumulating $\mathbf{J}_n,$ from all frames, we form the final \textit{Joint Cloud}:
$\mathcal{J} = \left\{\mathbf{J}_1, \mathbf{J}_2, \cdots, \mathbf{J}_N\right\}$.

As analyzed, $\mathcal{J}$ is extremely noisy and redundant. To reduce computational complexity, we establish a simple algorithm to cluster \red{$\mathcal{J}$} into smaller ones based on spatial proximity. We first select joints that belong to the middle-hip type. Then we apply the K-Means algorithm to get each target body's centre. We separate joint candidates by calculating their Euler distance with each body centre with a predefined threshold. Algorithm.~\ref{alg:Framwork} describes the detailed Joint Cloud construction procedure. 

\subsection{Kinematic Structures from Trajectile and Skeletal token cross-embedding}\label{sec:enc}

Joint Cloud $\mathcal{J}$ consists of 3D candidates $j_{n,\dot{\mathbf{v}}}^{i,\dot{\mathbf{m}}}$ that are extremely noisy and redundant, especially with respect to their target ID and joint type labels.
In this work, we propose a Joint Cloud Selection and Aggregation Transformer (JCSAT) framework, which is a Transformer-based structure as shown in Fig.~\ref{fig:system}. 
Inspired by DMAE~\cite{jiang2022dual}, 
each 3D candidate $j_{n,\dot{\mathbf{v}}}^{i,\dot{\mathbf{m}}}$ will be treated as a Transformer's token. JCSAT focuses on \textit{selecting} and \textit{aggregating} valid \textit{trajectile}, \textit{skeletal} and \textit{angular} features from noisy evidence. 
Specifically, JCSAT consists of three cascaded encoder modules \red{that} deeply explore the correlations among the candidate tokens along different dimensions. Each encoder is followed by an Optimal Token Attention Path (OTAP) module \red{which} differentiates and optimally selects reliable and representative candidate token features in a differentiable manner.
Finally, an MLP prediction head regresses the aggregated tokens to 3D human motion trajectories.

The input sequence (after padding),
$\mathcal{J} = \left\{\mathbf{J}_1, \mathbf{J}_2, \cdots, \mathbf{J}_N\right\}$,
can be organized as a \red{five-}dimensional matrix $\mathcal{J} \in \mathbb{R}^{N \times \dot{V} \times \dot{M} \times  I \times 3}$ 
where each dimension represents the frame index $n$, triangulation view pair index $\dot{\mathbf{v}}$, candidate index $\dot{\mathbf{m}}$, joint type $i$, and its corresponding 3D coordinate, respectively. Here $\dot{\mathbf{V}}$ and $\dot{\mathbf{M}}$ denote the total number of view combinations and candidate count (ID combination).
First, a trainable linear projection layer is used to embed each candidate $j_{n,\dot{\mathbf{v}}}^{i,\dot{\mathbf{m}}}$ into a high-dimensional feature. 
Then we rearrange the input features into \textit{Frame-Target-major} and \textit{Type-Target-major} data structures. In \textit{Frame-Target-major} data structure, we encode all tokens from the same view combination and the same joint type by \textit{Trajectory Encoder} ($T_{\mathrm{Enc}}$). In \textit{Type-Target-major} data structures, similarly, we encode all tokens from the same view combination and the same frame index by \textit{Structure Encoder} ($S_{\mathrm{Enc}}$). These two procedures can be formulated as:
\begin{equation}
\label{eq.traj}
\begin{split}
\hat{z}_{T}^{i,\dot{\mathbf{v}}} & = \mathit{T}_{\mathrm{OTAP}} (\mathit{T}_{\mathrm{Enc}}(\mathit{proj}(\{j_{n,\dot{\mathbf{v}}}^{i,\dot{\mathbf{m}}}~|~for ~\mathbf{all}~n~\mathbf{and}~\dot{\mathbf{m}}\})
\\ & +Pos_{T})) \\
& =\left \{ z_{0, \dot{\mathbf{v}}}^{i}, z_{1, \dot{\mathbf{v}}}^{i}, \cdots , z_{n, \dot{\mathbf{v}}}^{i} \right \} ,
\end{split}
\end{equation}
\begin{equation}
\label{eq.stru}
\begin{split}
\hat{z}_{S}^{n,\dot{\mathbf{v}}} & = \mathit{S}_{\mathrm{OTAP}} (\mathit{S}_{\mathrm{Enc}}(\mathit{proj}(\{j_{n,\dot{\mathbf{v}}}^{i,\dot{\mathbf{m}}}~|~for ~\mathbf{all}~i~\mathbf{and}~\dot{\mathbf{m}}\})
\\ & +Pos_{S})) \\
& = \left \{ z^0_{n,\dot{\mathbf{v}}}, z^1_{n,\dot{\mathbf{v}}}, \cdots , z^i_{n,\dot{\mathbf{v}}} \right \} ,
\end{split}
\end{equation}
where $\hat{z}$ is a set of the encoded joint patch, $\mathit{proj}$ is the projection layer, and $Pos$ is the position encoding. To be more precise, $\hat{z}_{T}^{i,\dot{\mathbf{v}}}$ and $\hat{z}_{S}^{n,\dot{\mathbf{v}}}$ are designed to represent $i$-th joint's trajectory and $n$-th frame's body structure at $\dot{\mathbf{v}}$-th view pair respectively. Each OTAP module aggregates the features by reducing the data on the candidate dimension $\dot{\mathbf{m}}$. In other words, OTAP helps us to select the most suitable representatives in terms of the joint's trajectory and the body's structure.
Next, we blend two representations by element-wise addition as Trajectile and Skeletal cross-embedding to share spatial and temporal information cohesively. After that, the redundant view dimension is encoded by $V_{\mathrm{Enc}}$ and aggregated by the following OTAP module:
\begin{equation}
\label{eq.view}
\begin{split}
\hat{z}_{V}^{i,n} & = \mathit{V}_{\mathrm{OTAP}}(\mathit{V}_{\mathrm{Enc}}(\mathit{proj}(\left \{ \hat{z}_{T}^0+\hat{z}_{S}^0, \cdots , \hat{z}_{T}^{\dot{\mathbf{v}}}+\hat{z}_{S}^{\dot{\mathbf{v}}} \right \}) 
\\ & + Pos_{V})) ,
\end{split}
\end{equation}
where $\hat{z}_{V}^{i,n}$ represents the $i$-th joint's token at $n$-th frame. It can be regarded as a motion representation. Last, an MLP prediction head is used to regress the motion representation to 3D movements.

\subsubsection{Position Encoding}

Similar to the vanilla Transformer~\cite{vaswani2017attention}, we add position encoding to facilitate the Transformer in distinguishing the input tokens. Following DMAE\cite{jiang2022dual}, we employ the Fourier encoding~\cite{tancik2020fourier} to encode the joint position, frame index, joint type index and view index \red{respectively}:
\begin{equation}
PosEnc(x) = \left [ \textrm{cos}(2\pi \mathbf{B} x),\textrm{sin}(2\pi \mathbf{B} x) \right ]^T ,
\end{equation}
where $x \in \mathbb{R}^{D_{emb}\times D_{in}} $ is the input, and $\mathbf{B}\in\mathbb{R}^{D_{out}\times D_{emb}}$ is a random matrix in Gaussian distribution which is sampled from $\mathcal{N}(0,\sigma ^2)$ and $\sigma$ is a hyperparameter. $D_{emb}$, $D_{in}$ and $D_{out}$ denote the dimension of embedding, input and output respectively. In this way, the $Pos_{T}$, $Pos_{S}$ and $Pos_{V}$ in Eq.~\ref{eq.traj}, \ref{eq.stru}, \ref{eq.view} can be formulated as:
\begin{gather}
Pos_{T} = PosEnc(j_{n,\dot{\mathbf{v}}}^{i,\dot{\mathbf{m}}}) + PosEnc(n) , \\
Pos_{S} = PosEnc(j_{n,\dot{\mathbf{v}}}^{i,\dot{\mathbf{m}}}) + PosEnc(i) , \\ 
Pos_{V} = PosEnc(\dot{\mathbf{v}}) .
\end{gather}

\subsubsection{Joint type conversion in MLP}

The skeleton type is inconsistent between the 2D detector and the 3D ground truth. The 2D detector we use predicts 25-joint skeletons named \textit{Body25} while in the Shelf or Campus dataset, it provides 14-joint skeletons named \textit{Shelf14}. Existing works~\cite{zhang20204d,dong2021fast,lin2021multi} rely on hard-code interpolation to convert the reconstructed skeleton format to the ground-truth skeleton format for training or evaluation. Jiang et al.~\cite{jiang2022dual} adopt the MAE~\cite{he2022masked} concept to complete \textit{Shelf14} with \textit{Body25} during inference. However, both conversion approaches have their drawbacks. The hard-code interpolation discards several joints, such as the foot toe and only keeps the ankle while the toe's location does affect the ankle's prediction. DMAE~\cite{jiang2022dual}'s implicit completion leads to expensive computation. Different from them, we propose to use a simple and lightweight MLP layer to convert the skeleton format. Specifically, we fully connect the latent code on the joint type dimension and reduce the joint type number from $I$ to $I'$.

\subsubsection{Joint Cloud Masking}\label{sec:mask} 

Each view pair could observe different body targets, leading to inconsistent candidate numbers. To rectify this, Joint Cloud is padded with zero to maintain a fixed-dimensional matrix form. Naturally, the zero padding should be masked in the self-attention mechanism~\cite{vaswani2017attention}. We also perform token masking during training, to mimic the masked language modelling task~\cite{devlin2018bert}. To increase the model's robustness, we introduce randomness by randomly masking the input candidates: randomly select a mask ratio and apply it in every training batch.

\subsection{Differentiable Feature Selection and Aggregation}\label{sec:jcsa}

Considerable noise and redundancies exist in the output of the cross-embedding encoders $T_{\mathrm{Enc}}$, $S_{\mathrm{Enc}}$ and $V_{\mathrm{Enc}}$. As shown in Fig.~\ref{fig:system}, $T_{\mathrm{Enc}}$ encodes $N \times \dot{M}$ candidate tokens while only $N$ tokens are regarded as output representing trajectory details for each joint. $S_{\mathrm{Enc}}$ encodes $I \times \dot{M}$ candidate tokens while only $I$ tokens are regarded as output representing the skeletal structure of the current frame. And $V_{\mathrm{Enc}}$ encodes $(I \times N) \times \dot{V}$ tokens while only $(I \times N)$ tokens represent the cross-view information. As such, our aim is to reduce redundancy from $\mathcal{J}$ and calculate representative token features in a differentiable manner. 

One straightforward idea is to directly aggregate information from all candidate tokens and rely on the Transformer's self-attention mechanism to differentiate useful information from noise. This is in a way similar to ViT~\cite{dosovitskiy2020image} which aggregates the class token from image patches. We refer to it as \textit{Non-selective aggregation}. However, we argue that the 3D joint candidates from $\mathcal{J}$ do not conform to any rigorous statistical distribution, i.e., the discrepancies existing between the candidates are mutually exclusive in the sense that there should be one that is most correct, while all the others will only contribute negatively to the final regression accuracy when considered inclusively. We will validate this assumption in the ablation study.

As such, we propose an Optimal Token Attention Path (OTAP) module to select the most representative token(s) along desirable dimensions in a differentiable manner, named \textit{selective aggregation}. Given latent representation in the form $z \in \mathbb{R}^{K_1\times K_2\times K_3}$, and a loss function $\mathcal{L}_{\mathrm{OTAP}}(z)$. Our goal is to minimize the loss function by selecting only one candidate along the second dimension $k_2$, thereby arriving at a more compact representation $z' \in \mathbb{R}^{K_1\times K_3}$ in the end. 
Such selection task can be considered as an optimal transportation problem~\cite{mialon2020trainable}, 
in which the transportation plan $\mathbf{P}\in\mathbb{R}^{ (K_1 \times K_2) \times K_3}$ converts embedding to $z'$:
\begin{equation}
z'=\sqrt{K_2} \times \left \{ \sum_{k=1}^{K_1\times K_2}\mathbf{P}_{k,1} \bar{z}_k, \cdots, \sum_{k=1}^{K_1 \times K_2}\mathbf{P}_{k,K_2} \bar{z}_k\right \} ,
\end{equation}
where $\bar{z}\in\mathbb{R}^{(K_1\times K_2)\times K_3}$ is calculated from $z \in \mathbb{R}^{K_1\times K_2\times K_3}$ by vectorizing along its first and second dimensions, and $\bar{z}_k\in\mathbb{R}^{1\times K_3}$ denotes the $k$-th row of $\bar{z}$. We ensure $\mathbf{P}$ approximates a permutation matrix, i.e., having the row sum and column sum close to 1, via regularization with Sinkhorn's algorithm~\cite{sinkhorn1967diagonal}.

The OTAP module will be used to select optimal tokens along the joint candidate dimension ($k_2= \dot{\mathbf{m}}$) for $T_{\mathrm{Enc}}$ and $S_{\mathrm{Enc}}$. However, for $T_{\mathrm{Enc}}$, $k_1$ dimension will be set for the frame indices $n$, while for $S_{\mathrm{Enc}}$, $k_1$ will be set for join type $i$, respectively.
For the view encoder $V_{\mathrm{Enc}}$, OTAP will be employed to select the most informative viewing angle ($k_2= \dot{\mathbf{v}}$), and the $k_1$ dimension will be set for $i\times n$.

\textbf{Remark 1.} We want to make sure that the token selection process should be done in a differentiable manner, since we will need to progressively \textit{reconsider}, and \textit{update} the selection outcomes after checking on trajectory consistency, skeletal structural and angular factors in an iterative manner. And the selection process will in-turn help to improve the cross-embedding encoders to generate more discriminative features among the noisy observations. 

\textbf{Remark 2.} When we remove the proposed OTAP module, the aggregation process will be downgraded to non-selective aggregation. The merit of the OTAP module is evaluated in Table~\ref{table:ab_aggregate}.

\subsection{Loss functions}

Several loss functions are carefully designed to fulfil the end-to-end learning manner.
First, we use Mean Squared Error (MSE) to compute the reconstruction loss $\mathcal{L}_{rec}$ between the prediction and ground-truth 3D joints in Cartesian space:
\begin{equation}\label{eq:l_rec}
\mathcal{L}_{rec}=\frac{1}{N + I}\sum_{n=1}^{N}\sum_{i=1}^{I}\left\| Pred_i^n-Gt_i^n\right\| ,
\end{equation}
where $Pred_i^n$ and $Gt_i^n$ represent the prediction and ground-truth 3D joint $i$ at frame $n$, respectively. $\left\| \cdot \right\|$ represents the L2 norm.

Second, we project both prediction $Pred_i^n$ and ground-truth $Gt_i^n$ to 2D to compute each view's projection loss $\mathcal{L}_{proj}$:
\begin{equation}\label{eq:l_proj}
\mathcal{L}_{proj}=\sum_{v=1}^{V}(\frac{1}{N + I}\sum_{n=1}^{N}\sum_{i=1}^{I}\left\| \prod_{v}Pred_i^n-\prod_{v}Gt_i^n\right\|) ,
\end{equation}
where $\prod_v$ is $v$-th camera's projection matrix.

Third, for kinematic structure learning, we adopt one of the 3D HPE evaluation metrics, the Percentage of Correctly estimated Parts (PCP) as the Kinematic Loss $\mathcal{L}_{pcp}$:
\begin{equation}\label{eq:l_pcp}
\begin{split}
\mathcal{L}_{pcp}=\frac{1}{N+L}\sum_{n=1}^{N}\sum_{\left \{ a,b \right \}\in L }^{L}Relu(\left\| Pred_a^n-Gt_a^n\right\| \\ 
+ \left\| Pred_b^n-Gt_b^n\right\| - \left\| Gt_a^n-Gt_b^n\right\|) ,
\end{split}
\end{equation}
where $L$ represents the limb between joint $a$ and joint $b$. 
$Relu$ in this equation \red{acts} as a hinge loss
which enables $\mathcal{L}_{pcp}$ equal to 0 when the reconstructed limb is correct (See Sec.~\ref{sec:eval_metrics} for details). We also compute the projected Kinematic Loss.
Moreover, the human body is nearly symmetric and consistent during movement. That means the right side's bone length should be similar to the left side's. The length of the same bone should be consistent across frames. Based on this observation, we compute the structure loss $\mathcal{L}_{bone}$ to enable the above constraint:
\begin{equation}\label{eq:l_bone}
\begin{split}
\mathcal{L}_{bone}=\frac{1}{N+L-1}
\sum_{n=2}^{N}
\sum_{l=1}^{L}
\left\| Limb_l^{n-1}-Limb_l^{n}\right\| \\
+ \frac{1}{N+S}
\sum_{n=2}^{N}
\sum_{\left \{ r,l\right \} \in S }^{S}
\left\| Limb_r^{n}-Limb_l^{n}\right\| ,
\end{split}
\end{equation}
where $Limb$ represents the predicted limb length and $l$ denotes the limb index. $\left \{ r,l\right \} \in S$ represents the symmetric limb pair. In practice, we only compute the upper/lower arms, upper/lower legs and right/left eye-to-ear limbs, eye-to-nose limbs in OpenPose~\cite{cao2019openpose} body format.
Overall, the final loss $\mathcal{L}$ is given by:
\begin{equation}
\mathcal{L}=\mathcal{L}_{rec}+\mathcal{L}_{proj}+\mathcal{L}_{pcp}+\mathcal{L}_{bone}.
\end{equation}

\section{Experiments}

\subsection{Experimental Setup}

\subsubsection{Dataset}

\red{We evaluate our framework against state-of-the-art methods using the following datasets. The limited availability of annotated multi-view multi-person motion datasets highlights the necessity and value of our BUMocap and BUMocap-X.}
\begin{enumerate} [label=(\roman*)]
\item The \textbf{Campus} and the \textbf{Shelf} datasets. These are two \red{widely used benchmarks proposed by} \cite{belagiannis20143d,belagiannis2014multiple} where the Campus captures the outdoor activities of three individuals over three calibrated cameras, while the Shelf captures four individuals with five cameras indoors. Two datasets have
annotated 1231 and 831 frames, respectively.
However, these frames are partially labelled, leading to few training data.
\item The \textbf{BUMocap}~\cite{jiang2022dual} and \textbf{BUMocap-X} datasets. The BUMocap records 90 seconds of challenging 3-actor interactive motions from five calibrated cameras \red{with fully manual annotation}. In order to better evaluate the performance of various methods under \textit{serious occlusion}, we further \red{expand it and }present \textbf{BUMocap-X}, with an additional 120 seconds of extremely challenging 5-actor interactive motion from five cameras. \red{BUMocap-X records group dance routines including self-rotations and positional changes. Given the nature of the dance movements, we believe that the richness of the actions in this dataset far surpasses that of the previous one.}
\item The \textbf{CMU Panoptic} dataset~\cite{joo2015panoptic}. It is currently the largest dataset with 65 sequences, and 1.5 million frames of 3D skeletons under 8 motion categories. The annotation is auto-generated by Kinect~\cite{azurekinect} build-in algorithms. Note that most individuals are without occlusion, which makes pose estimation relatively easy. Following ~\cite{reddy2021tessetrack}, we select data from five cameras (i.e., HD cameras 3, 6, 12, 13, 23) for evaluation.
\item The \textbf{Markerless} dataset~\cite{zhang20204d}. It consists of 4 video sequences with challenging 6-actor interactive motions. Since there is no ground truth, we use this only for qualitative evaluation.
\end{enumerate}


\subsubsection{Evaluation Metrics}\label{sec:eval_metrics}

Following the evaluation protocols in previous works~\cite{belagiannis20153d,zhang20204d,dong2021fast,reddy2021tessetrack,lin2021multi,wu2021graph,jiang2022dual}, we first used the Percentage of Correctly estimated Parts (PCP) for evaluation. PCP indicates the correctness of the reconstructed limb by comparing the ground-truth limb length and the distance sum between reconstruction and ground truth at the joint level. Moreover, 
we used the Mean Per Joint Position Error (MPJPE) and Percentage of Correctly estimated Keypoints (PCK) as additional metrics. 
MPJPE measures the total average distance for the whole skeleton.
PCK measures the true joint if the distance between the reconstruction and the ground truth joints is smaller than a certain threshold. Following 4DAssocGraph~\cite{zhang20204d}, we set the threshold at 0.2 meters to calculate the Precision-Recall metric. 

\subsubsection{Implementation Details}
We implemented our model with three standard ViTs~\cite{dosovitskiy2020image}. $T_{\mathrm{Enc}}$ and $S_{\mathrm{Enc}}$ share the same structure. Similar to \cite{he2022masked}'s asymmetrical structure design, we make $V_{\mathrm{Enc}}$ 50\% narrower and shallower than $T_{\mathrm{Enc}}$ and $S_{\mathrm{Enc}}$ for reducing the computing cost. We adopted \cite{mialon2020trainable}'s optimal transport implementation in the OTAP module because of their open source well-managed package.
During inference, we apply a shifting window to gather candidates from the Joint Cloud. We set the window length as $N=10$. 

For the Shelf dataset, we followed previous works~\cite{belagiannis20143d,belagiannis2014multiple,belagiannis20153d} in training and evaluating configurations. In addition, We synthesized the ground truth for those non-labelled or non-complete-labelled frames by \cite{zhang20204d}. For the Campus dataset, we fine-tuned the model trained on the Shelf for 50 epochs. Following \cite{dong2021fast,reddy2021tessetrack,lin2021multi}, we adopted HRNet~\cite{sun2019deep} as the 2D detector for fairness of comparison. Since HRNet predicts \textit{Coco17}~\cite{lin2014microsoft} skeletons, we padded the 2D detection to be consistent with \textit{Body25}~\cite{cao2019openpose}.
For the training scheme, we used a dropout rate of 0.1. We adopted the AdamW optimizer with a cosine decay beginning with $10^{-5}$. The training epoch was set to 120. The batch size was set to 5. 
To \red{ease} the problem of limited training data, we performed orientation-based data augmentation as follows: (i) every input motion sequence was subtracted by the cluster centre which belongs to the first frame of the sequence; (ii) randomly rotated Joint Cloud along the vertical direction. More details can be found in the supplementary material.

\setlength{\tabcolsep}{3pt}
\begin{table}[t]
\begin{center}
\caption{Quantitative comparison on the Shelf dataset. PCP (\%) is used as the evaluation metric. ``AVG'' means the average PCP (\%) score of three actors (A1-A3). ``$\dagger$'' indicates \red{recalculated performance values for this experiment setup, which could differ from the original papers}.}
\label{table:shelf_PCP}
\begin{tabular}{rccccl}
\toprule
Method & A1 & A2 & A3 & AVG & Pub. Venue \\
\midrule
3DPS~\cite{belagiannis20153d} & 75.3 & 69.7 & 87.6 & 77.5 & TPAMI'15 \\
4DAssocGraph~\cite{zhang20204d} & 99.0 & 96.2 & 97.6 & 97.6 &  CVPR'20\\
MVPose~\cite{dong2021fast} & 98.8 & 94.1 & 97.8 & 96.9 &  TPAMI'21\\
TesseTrack~\cite{reddy2021tessetrack} & 99.1 & 96.3 & 98.3 & 97.9$\dagger$ &  CVPR'21\\
PlaneSweepPose~\cite{lin2021multi} & 99.3 & 96.5 & 98.0 & 97.9 & CVPR'21\\
MMG-CRG~\cite{wu2021graph} & 99.3 & 96.5 & 97.3 & 97.7 & ICCV'21 \\
DMAE~\cite{jiang2022dual} & \textbf{99.7} & 94.1 & \textbf{98.4} & 97.4 &  MM'22 \\
QuickPose~\cite{zhou2022quickpose} & 99.5 & 96.7 & 98.2 & 98.1 &  SIGGRAPH'22 \\
\red{FasterVoxelPose~\cite{ye2022faster}} & 99.4 & 96.0 & 97.5 & 97.6 & ECCV'22 \\
\red{TEMPO~\cite{choudhury2023tempo}} & 99.0 & 96.3 & 98.2 & 97.8$\dagger$ & ICCV'23 \\
\red{SelfPose3D~\cite{srivastav2024selfpose3d}} & 97.2 & 90.3 & 97.9 & 95.1 & CVPR'24\\
\midrule
Ours & 99.3 & \textbf{97.0} & 98.2 & \textbf{98.2} \\
\bottomrule
\end{tabular}
\end{center}
\end{table}

\setlength{\tabcolsep}{3pt}
\begin{table}[t]
\begin{center}
\caption{Quantitative comparison on the Campus dataset. PCP (\%) is used as the evaluation metric. ``$\dagger$'' indicates \red{recalculated performance values for this experiment setup, which could differ from the original papers}. ``$\ddagger$'' indicates \red{recalculated values based on our own re-implementation.}}
\label{table:campus_PCP}
\begin{tabular}{rccccl}
\toprule
Method & A1 & A2 & A3 & AVG & Pub. Venue \\
\midrule
3DPS~\cite{belagiannis20153d} & 93.5 & 75.7 & 84.4 & 84.5 & TPAMI'15\\
4DAssocGraph$\ddagger$~\cite{zhang20204d} & 64.8 & 82.0 & 96.6 & 81.1 & CVPR'20\\
MVPose~\cite{dong2021fast} & 97.6 & 93.3 & 98.0 & 96.3 & TPAMI'21\\
TesseTrack~\cite{reddy2021tessetrack} & 97.9 & 95.2 & \textbf{99.1} & \textbf{97.4} & CVPR'21\\
PlaneSweepPose~\cite{lin2021multi} & 98.4 & 93.7 & 99.0 & 97.0 & CVPR'21\\
\red{FasterVoxelPose~\cite{ye2022faster}} & 96.5 & 94.1 & 97.9 & 96.2 & ECCV'22\\
\red{TEMPO~\cite{choudhury2023tempo}} & 97.7 & \textbf{95.5} & 97.9 & 97.0$\dagger$ & ICCV'23\\
\red{SelfPose3D~\cite{srivastav2024selfpose3d}} & 92.5 & 82.2 & 89.2 & 87.9 & CVPR'24\\
\midrule
Ours & \textbf{99.6} & 93.5 & 98.8 & 97.3 \\
\bottomrule
\end{tabular}
\end{center}
\end{table}

\subsection{Quantitative Evaluation}

For the Shelf dataset, quantitative evaluations using the PCP metric are presented in Table~\ref{table:shelf_PCP}. We compared our method with recent state-of-the-art 3D HPE methods which can be generally divided into two categories: (1) optimization-based approaches~\cite{zhang20204d, dong2021fast, zhou2022quickpose, belagiannis20153d} and (2) learning-based approaches~\cite{reddy2021tessetrack, lin2021multi, jiang2022dual, wu2021graph,choudhury2023tempo,ye2022faster,srivastav2024selfpose3d}. Note that for the results presented in Table~\ref{table:shelf_PCP}, only \textbf{Actor-2 is heavily occluded}, \red{which substantially increases the difficulty of joint location reconstruction}. It can be seen that our method showed more obvious advantages compared to other methods in this challenging scenario.

For the Campus dataset, results are presented in Table~\ref{table:campus_PCP}. We compared the performances among three optimization-based approaches~\cite{dong2021fast, zhang20204d, belagiannis20153d} and five learning-based approaches~\cite{reddy2021tessetrack, lin2021multi,choudhury2023tempo,ye2022faster,srivastav2024selfpose3d}. We observed that there exists a large PCP score gap between the optimization-based and learning-based approaches. We believe this gap is caused by imprecise camera calibration, which leads to inaccurate triangulation. Since optimization-based methods are more heavily reliant on these initial triangulation and camera association, their performance on the Campus dataset was seriously affected/degraded. Moreover, our method learnt the moving trajectory from Actor-1 and Actor-3 while Actor-2 is stationary most of the time in the video sequence, which decreases the performance. 

Table~\ref{table:x_eval_2} demonstrates \red{the performance of state-of-the-art methods~\cite{zhang20204d,jiang2022dual,ye2022faster,choudhury2023tempo}} on the BUMocap and BUMocap-X datasets in terms of PCP, Precision, Recall and MPJPE metrics. Following \cite{jiang2022dual}, we trained our model and fine-tuned it on BUMocap and BUMocap-X datasets. \red{We trained Faster VoxelPose~\cite{ye2022faster} and TEMPO~\cite{choudhury2023tempo} under the same configurations mentioned in their papers respectively.} We unified those skeletons in \textit{Skel19}~\cite{zhang20204d} instead of \textit{Coco17}~\cite{lin2014microsoft} used in \cite{jiang2022dual}, in order to prevent unfair comparison caused by false skeleton conversion when interpolating joints are missing. As shown in Table~\ref{table:x_eval_2}, our method \textit{significantly} outperforms other methods. We've noticed that DMAE~\cite{jiang2022dual} failed to reconstruct most of the movements for the BUMocap-X dataset. We believe it failed because the simple re-ID algorithm in DMAE~\cite{jiang2022dual} failed to distinguish individuals with similar outlooks. 
\red{Faster VoxelPose~\cite{ye2022faster} didn't perform well for both datasets. We believe the reason lies that it cannot model the connection between voxelized joint heatmaps, especially when two people overlap causing mutual ambiguities. TEMPO~\cite{choudhury2023tempo} utilizes temporal information to ease the above issue, thus the performance is higher than \cite{ye2022faster}. Although TEMPO~\cite{choudhury2023tempo} proposes to utilize Spatial GRUs to model pose features in the temporal domain, it failed behind us, because our model explores the correlation among poses as well as joints, making our model able to resolve ID/joint ambiguities.}
In addition, MVPose~\cite{dong2021fast} crashed frequently due to re-ID failure. Therefore, we didn't present its quantitative evaluation, while qualitative demonstrations are provided in the supplementary material.

For the CMU Panoptic dataset~\cite{joo2015panoptic}, evaluation results are shown in Table~\ref{table:panoptic_eval}. As shown in the table, we found that our proposed method outperforms others by a small margin. In fact, all methods present relatively high-performance scores due to the fact that the CMU Panoptic dataset presents less challenging occlusion problems. 

\setlength{\tabcolsep}{1.5pt}
\begin{table}[t]
\begin{center}
\caption{Quantitative comparison on the BUMocap (BU) and BUMocap-X (BU-X) datasets. Average PCP(\%), Precision (\%), Recall (\%) and MPJPE (mm) are reported. ``$\ddagger$'' indicates \red{recalculated values based on our own re-implementation.}}
\label{table:x_eval_2}
\begin{tabular}{rcccc}
\toprule
Method & PCP $\uparrow$ & Precision $\uparrow$ & Recall $\uparrow$ & MPJPE $\downarrow$ \\
Dataset & BU/BU-X & BU/BU-X & BU/BU-X & BU/BU-X \\
\midrule
4DAssocGraph~\cite{zhang20204d} & 82.1/73.1 & 95.1/95.1 & 94.4/95.1 & 70.8/56.5 \\
DMAE~\cite{jiang2022dual} & 94.3/57.0 & 97.5/77.5 & 97.5/48.4 & 56.3/113.4 \\
\red{FasterVoxelPose~\cite{ye2022faster}} & 82.0/74.8 & 86.2/67.5 &  86.2/67.5 & 105.7/163.0 \\
\red{TEMPO$\ddagger$~\cite{choudhury2023tempo}} & 88.7/81.8 & 86.4/77.1 & 86.4/77.1 & 96.5/136.5 \\
\midrule
Ours & \textbf{95.5}/\textbf{88.5} & \textbf{97.9}/\textbf{97.8} & \textbf{97.9}/\textbf{97.8} & \textbf{52.2}/\textbf{49.7} \\
\bottomrule
\end{tabular}
\end{center}
\end{table}

\setlength{\tabcolsep}{4pt}
\begin{table}[t]
\begin{center}
\caption{Quantitative comparison with \cite{zhang20204d} on the CMU Panoptic dataset. We report our performance in terms of the detailed PCP (\%), average PCP (\%) and average MPJPE (mm). ``LU.'', ``RU.'', ``LL.'' and ``RL.'' stand for ``Left Upper'', ``Right Upper'', ``Left Lower'' and ``Right Lower'' respectively.}
\label{table:panoptic_eval}
\begin{tabular}{lcccccc}
\toprule
Subset ID   & \multicolumn{2}{c}{160224}    & \multicolumn{2}{c}{160422}     & \multicolumn{2}{c}{160906} \\
Subset Name & \multicolumn{2}{c}{haggling1} & \multicolumn{2}{c}{ultimatum1} & \multicolumn{2}{c}{pizza1} \\
Body Num.   & \multicolumn{2}{c}{3}         & \multicolumn{2}{c}{7}          & \multicolumn{2}{c}{6}      \\ 
\midrule
Method    & Ours          & \cite{zhang20204d} & Ours          & \cite{zhang20204d} & Ours          & \cite{zhang20204d} \\ 
\midrule
LU. Arm   & \textbf{99.9} & \textbf{99.9} & \textbf{99.9} & \textbf{99.9} & \textbf{99.9} & \textbf{99.9} \\
RU. Arm   & \textbf{99.9} & \textbf{99.9} & \textbf{98.9} & 97.3          & 98.1          & \textbf{99.9} \\
LL. Arm   & \textbf{99.6} & 99.4          & \textbf{95.1} & 88.9          & \textbf{99.9} & \textbf{99.9} \\
RL. Arm   & \textbf{98.6} & 97.3          & \textbf{96.6} & 91.9          & 93.2          & \textbf{93.9} \\
LU. Leg   & \textbf{99.9} & \textbf{99.9} & \textbf{98.9} & \textbf{98.9} & \textbf{99.9} & 97.0          \\
RU. Leg   & \textbf{99.9} & 99.5          & 99.6          & \textbf{99.9} & 97.3          & \textbf{99.9} \\
LL. Leg   & \textbf{99.9} & \textbf{99.9} & 98.2          & \textbf{99.9} & \textbf{98.2} & 97.0          \\
RL. Leg   & \textbf{99.9} & \textbf{99.9} & 99.2          & \textbf{99.9} & \textbf{99.9} & \textbf{99.9} \\
Head      & 91.4          & \textbf{93.0} & 84.3          & \textbf{89.2} & 88.3          & \textbf{88.9} \\
Torso     & \textbf{99.9} & \textbf{99.9} & \textbf{99.9} & \textbf{99.9} & \textbf{99.9} & \textbf{99.9} \\ 
\midrule
AVG PCP $\uparrow$   & \textbf{98.9} & \textbf{98.9} & \textbf{97.1} & 96.6          & 97.5          & \textbf{97.6} \\
AVG MPJPE $\downarrow$ & \textbf{32.7} & 34.1          & \textbf{51.3} & 60.7          & 47.1          & \textbf{46.3} \\ 
\bottomrule
\end{tabular}
\end{center}
\end{table}

\begin{figure*}[tbhp]
\centering
\setlength{\abovecaptionskip}{-1.5mm}
\includegraphics[width=0.98\linewidth]{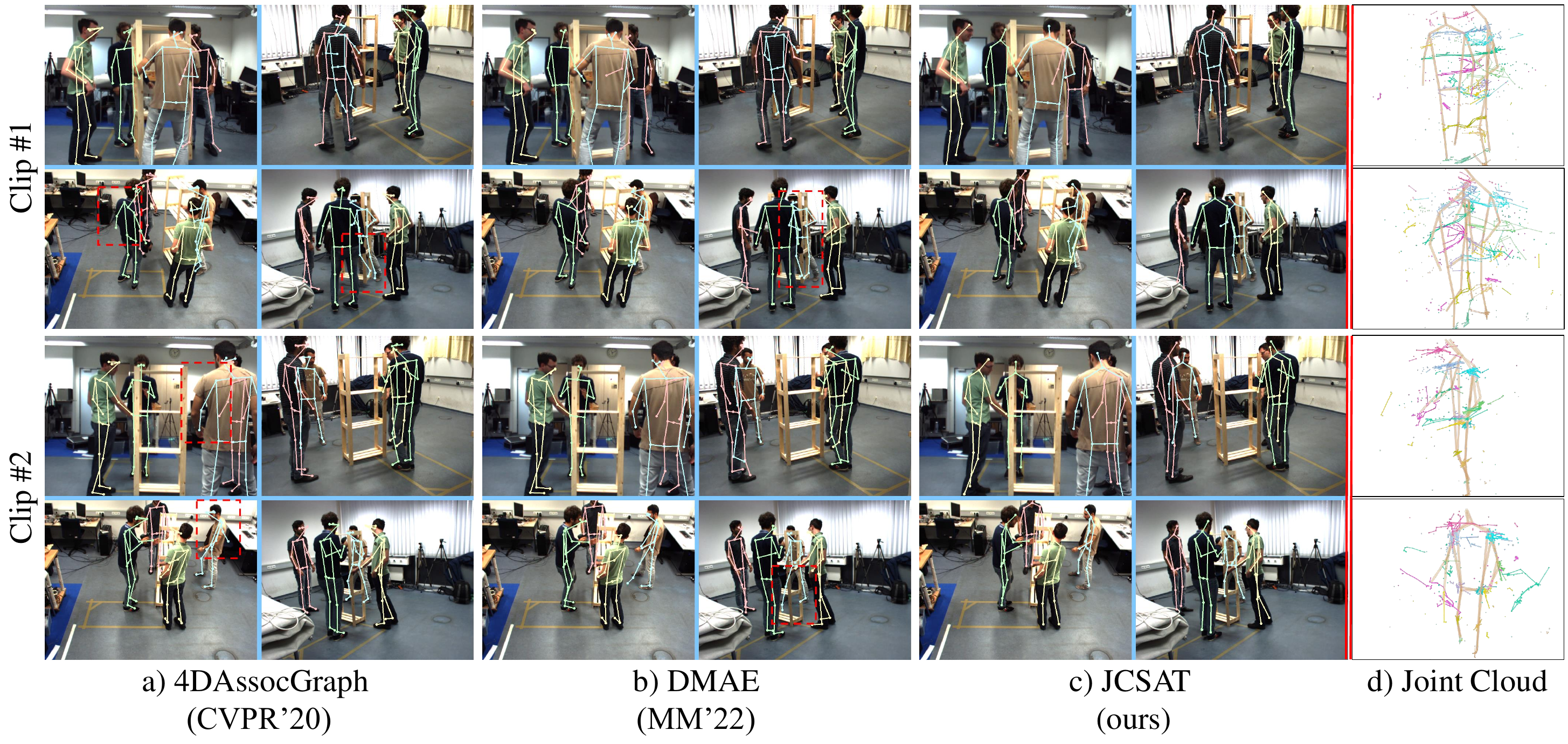}
\caption{Qualitative comparison with 4DAssocGraph~\cite{zhang20204d} and DMAE~\cite{jiang2022dual} on the Shelf dataset. We demonstrate each projection for two clips in four views. 
In addition, we demonstrate the Joint Cloud for Actor-2 across a sequence of 10 frames. For a clear view, all joints in the same type are connected to represent trajectories and the predicted skeleton is drawn as an indicator.}
\label{fig_shelf_out}
\end{figure*}

\begin{figure}[tbhp]
\centering
\setlength{\abovecaptionskip}{-1mm}
\includegraphics[width=0.98\linewidth]{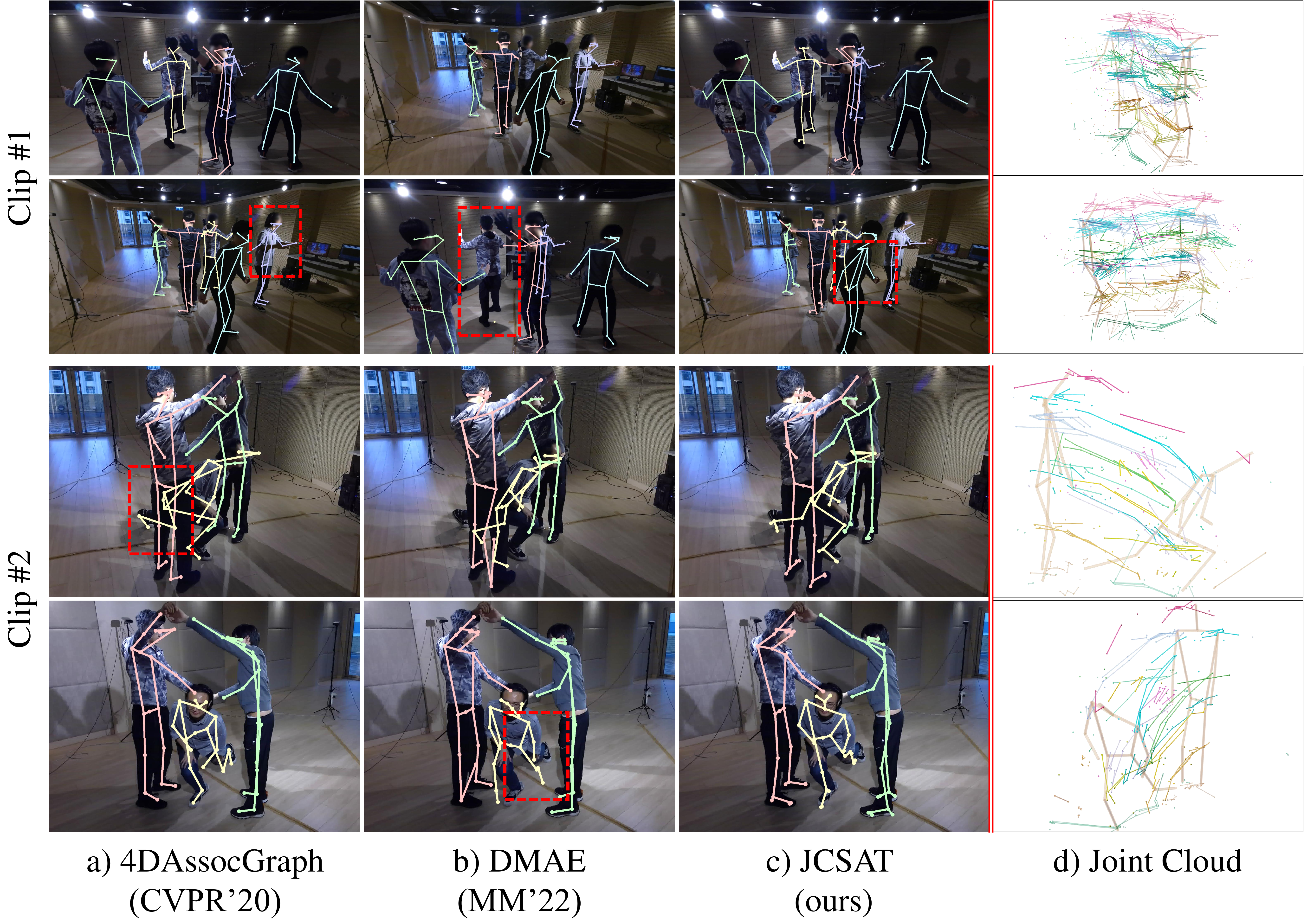}
\caption{Qualitative comparison with 4DAssocGraph~\cite{zhang20204d} and DMAE~\cite{jiang2022dual} on our newly collected dataset. We also visualize the Joint Cloud of Actor-1 and Actor-3 across 10 frames. The predicted skeletons from the first and last frames are used as indicators.
}
\label{fig_coop6}
\end{figure}

\begin{figure}[tbhp]
\centering
\setlength{\abovecaptionskip}{-1mm}
\includegraphics[width=0.98\linewidth]{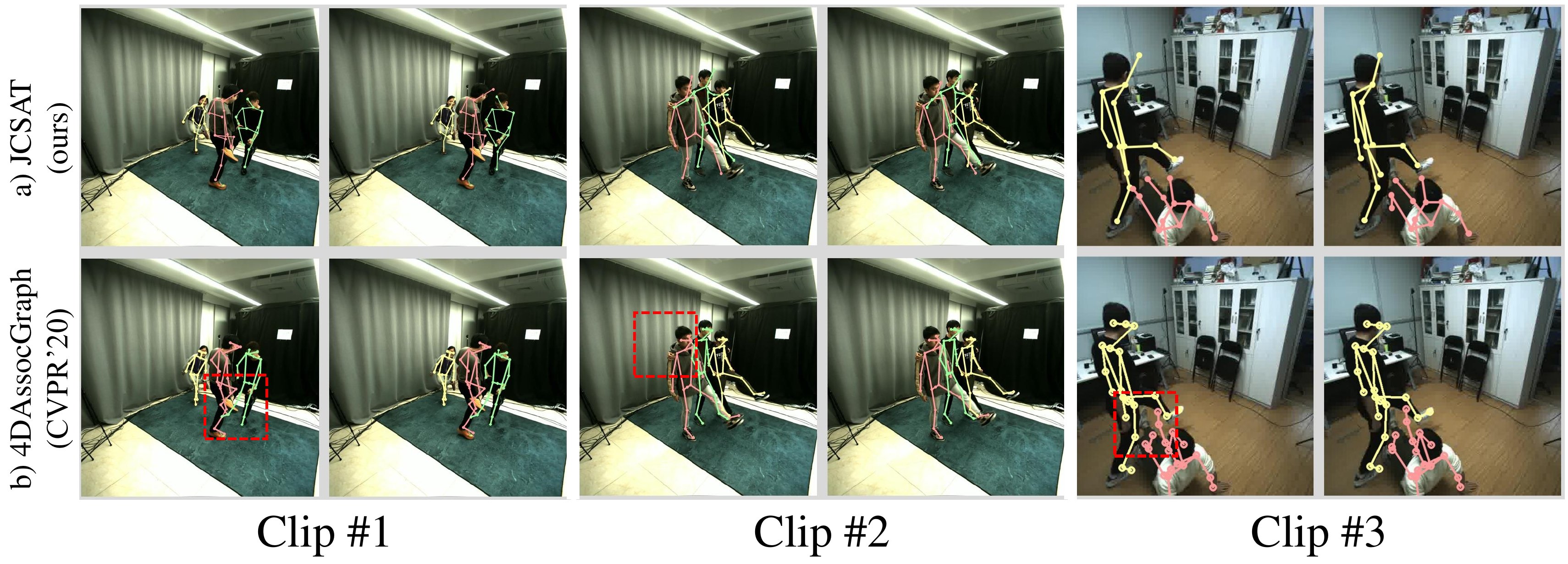}
\caption{Qualitative comparison with 4DAssocGraph~\cite{zhang20204d} on Markerless~\cite{zhang20204d} dataset. We visualize three video clips by comparing two consecutive frames to demonstrate the temporal consistency.
}
\label{fig_markerless}
\end{figure}

\subsection{Qualitative Evaluation}

The qualitative evaluation results on the Shelf, BUMocap, BUMocap-X, and the Markerless datasets are shown in 
Fig.~\ref{fig_shelf_out},
Fig.~\ref{fig_coop6}, Fig.~\ref{fig:teaser}, and Fig.~\ref{fig_markerless}, respectively. We provide visual comparisons of 4DAssocGraph~\cite{zhang20204d}, DMAE\cite{jiang2022dual}, TEMPO~\cite{choudhury2023tempo} and our proposed model. We highlight the falsely reconstructed parts with a red dotted box. 
As can be seen in Fig.~\ref{fig_shelf_out}, 4DAssocGraph~\cite{zhang20204d} estimated incorrect joint locations on the arms and legs under heavy occlusions, with wrong ID association. DMAE~\cite{jiang2022dual} predicted wrong foot locations because this prediction is related to other joints' temporal and structural information while other joints are associated similarly to \cite{zhang20204d}.  Our model predicted more reasonable and precise locations by using the trajectory and structural information for the unseen joints for Actor-2 in clip 2.
For the testing results over the Markerless dataset shown in Fig.~\ref{fig_markerless}, since we trained the models on the Shelf dataset and tested on the Markerless dataset, the skeleton format is \textit{Shelf14}, which differs from the output presented by 4DAssocGraph~\cite{zhang20204d}. Nevertheless, when considering the limbs, our proposed framework yields more faithful 3D movements.
More qualitative results can be found in the supplementary material.

\red{Moreover, to assess the generalization ability of our model, we conducted cross-dataset performance evaluations. Specifically, we trained the JCSAT using the Shelf, BUMocap and BUMocap-X datasets and then tested it on alternate datasets. Given that the Shelf dataset provides a 15-joint annotation, we standardized both the ground truth and the predictions to this skeleton definition. Other training or testing configurations remain the same. The results are presented in Table~\ref{table:ab_gen}. Our observations confirm that our model is able to transfer the learnt motion prior across different datasets. However, we noted a decline in performance when the model, trained on the Shelf dataset, was tested on the BUMocap or BUMocap-X datasets. We attribute this decrease in performance to the complexity and diversity of the test motions, which were not captured as extensively in the training motions from the Shelf dataset.}

\setlength{\tabcolsep}{8pt}
\begin{table}[t]
\begin{center}
\caption{Inference speed comparison with state-of-the-art works on the Shelf dataset.  We categorize these methods as optimization-based (\textit{Opt.}) and learning-based (\textit{Learn.}) methods. We report \red{their processing time in milliseconds per frame for the 2D-to-3D procedures. The average PCP scores are also provided.}}
\label{table:speed}
\begin{tabular}{rrcccc}
\toprule
Type & Method & PCP $\uparrow$ & Time$\downarrow$ \\ 
\midrule
\multirow{3}{*}{\textit{Opt.}} & 4DAssocGraph~\cite{zhang20204d}  & 97.6 & 31.9 \\
& MVPose~\cite{dong2021fast} & 96.9 & $\sim$40 \\
& QuickPose~\cite{zhou2022quickpose} & 98.1 & \textbf{2.94} \\ 
\midrule
\multirow{8}{*}{\textit{Learn.}} & TesseTrack~\cite{reddy2021tessetrack} & 97.9 & \textgreater{333} \\
& PlaneSweepPose~\cite{lin2021multi} & 97.9 & 23.4 \\
& MMG-CRG~\cite{wu2021graph} & 97.7 & $\sim$243 \\
& DMAE~\cite{jiang2022dual} & 97.4 & $\sim$78.5 \\ 
& \red{FasterVoxelPose~\cite{ye2022faster}} & 97.6 & 32.2 \\
\cmidrule(lr){2-4}
& Ours with 3 views & 96.4 & 76.5 \\
& Ours with 4 views & 97.5 & 117.4 \\
& Ours with 5 views & \textbf{98.2} & 229.1 \\
\bottomrule
\end{tabular}
\end{center}
\end{table}

\setlength{\tabcolsep}{6pt}
\begin{table}[t]
\begin{center}
\caption{Ablation study on the Shelf dataset. PCP (\%) is used as the evaluation metric. ``View'' indicates the number of views used during inference. ``\textit{mask}'' stands for Joint Cloud masking. ``\textit{kine. 
 loss}'' stands for the Kinematic Loss.}
\label{table:ab_pcp}
\begin{tabular}{cccccccc}
\toprule
\multicolumn{2}{c}{Model} & \multirow{2}{*}{View} & \multirow{2}{*}{A1} & \multirow{2}{*}{A2} & \multirow{2}{*}{A3} & \multirow{2}{*}{AVG} \\
\textit{mask} & \textit{kine. loss} &&&& \\
\midrule
\xmark & \xmark & 5 & 99.0 & 91.0 & 98.0 & 96.0 \\
\xmark & \cmark & 5 & 99.2 & 91.9 & 98.3 & 96.5 \\
\cmark & \xmark & 5 & 98.7 & 96.2 & \textbf{98.5} & 97.8 \\
\midrule
\cmark & \cmark & 2 & 95.6 & 86.8 & 92.0 & 91.5 \\
\cmark & \cmark & 3 & 98.2 & 94.1 & 97.1 & 96.4 \\
\cmark & \cmark & 4 & 99.1 & 95.2 & 98.1 & 97.5 \\
\midrule
\cmark & \cmark & 5 & \textbf{99.3} & \textbf{97.0} & 98.2 & \textbf{98.2} \\
\bottomrule
\end{tabular}
\end{center}
\end{table}

\setlength{\tabcolsep}{2pt}
\begin{table}[tbhp]
\begin{center}
\caption{Performance for JCSAT variants on the Shelf dataset. Average PCP(\%), Precision (\%), Recall (\%) and MPJPE (mm) are reported.}
\label{table:ab_aggregate}
\begin{tabular}{rcccc}
\toprule
Model & PCP $\uparrow$ & Precision $\uparrow$ & Recall $\uparrow$ & MPJPE $\downarrow$ \\
\midrule
OTAP-a & 96.9 & 99.4 & 99.4 & 59.2 \\  
OTAP-b & 96.4 & 99.0 & 99.0 & 58.0 \\  
OTAP-c & 97.3 & \textbf{99.5} & \textbf{99.5} & 57.4 \\  
\midrule
$T_{\mathrm{Enc}}$-S\&$S_{\mathrm{Enc}}$-S & 94.7 & 97.7 & 97.7 & 66.5 \\
$T_{\mathrm{Enc}}$-S\&$S_{\mathrm{Enc}}$-M & 96.5 & 99.0 & 99.0 & 63.5 \\
$T_{\mathrm{Enc}}$-M\&$S_{\mathrm{Enc}}$-S & 95.7 & 98.4 & 98.4 & 63.7 \\
$T_{\mathrm{Enc}}$-M\&$S_{\mathrm{Enc}}$-L & 97.1 & 99.3 & 99.3 & 62.5 \\
$T_{\mathrm{Enc}}$-L\&$S_{\mathrm{Enc}}$-M & 98.2 & 99.4 & 99.4 & 57.8 \\
$T_{\mathrm{Enc}}$-L\&$S_{\mathrm{Enc}}$-L & 98.1 & 99.6 & 99.6 & 58.6 \\
\midrule
$T_{\mathrm{Enc}}$-M\&$S_{\mathrm{Enc}}$-M & \textbf{98.2} & \textbf{99.5} & \textbf{99.5} & \textbf{57.3} \\
\bottomrule
\end{tabular}
\end{center}
\end{table}

\setlength{\tabcolsep}{2.5pt}
\begin{table}[t]
\begin{center}
\caption{Comparison with 4DAssocGraph~\cite{zhang20204d} (abbreviated as \textit{4DA}) on the Shelf dataset when camera parameters are inaccurate. PCP (\%) is used as the evaluation metric. ``Inac. views'' denotes the number of \textit{inaccurate views} during inference.}
\label{table:ab_incorrect_camera}
\begin{tabular}{rcccccccc}
\toprule
Inac. views & \multicolumn{2}{c}{A1} & \multicolumn{2}{c}{A2} & \multicolumn{2}{c}{A3} & \multicolumn{2}{c}{AVG} \\
Method & Ours & \textit{4DA} & Ours & \textit{4DA} & Ours & \textit{4DA} & Ours & \textit{4DA} \\
\midrule
1 & \textbf{99.8} & 98.9 & 93.5 & \textbf{95.7} & \textbf{98.2} & 97.1 & \textbf{97.2} & \textbf{97.2} \\
2 & \textbf{98.7} & 97.8 & \textbf{91.6} & 91.5 & \textbf{97.6} & \textbf{97.6} & \textbf{96.0} & 95.6 \\
3 & 45.2 & 44.1 & 29.2 & \textbf{37.5} & 30.6 & 31.4 & \textbf{35.0} & \textbf{37.7} \\
\midrule
0 & \textbf{99.3} & 99.0 & \textbf{97.0} & 96.2 & \textbf{98.2} & 97.6 & \textbf{98.2} & 97.6 \\
\bottomrule
\end{tabular}
\end{center}
\end{table}

\setlength{\tabcolsep}{2.5pt}
\begin{table}[t]
\begin{center}
\caption{Comparison with 4DAssocGraph~\cite{zhang20204d} (abbreviated as \textit{4DA}) on the Shelf dataset with different input image resolution rescaling factors. PCP (\%) is used as the evaluation metric.}
\label{table:ab_res}
\begin{tabular}{rcccccccc}
\toprule
Factor & \multicolumn{2}{c}{A1} & \multicolumn{2}{c}{A2} & \multicolumn{2}{c}{A3} & \multicolumn{2}{c}{AVG} \\
Method & Ours & \textit{4DA} & Ours & \textit{4DA} & Ours & \textit{4DA} & Ours & \textit{4DA} \\
\midrule
x0.25 & \textbf{99.7} & 98.7 & 94.1 & \textbf{95.3} & \textbf{98.4} & 97.7 & \textbf{97.4} & 97.2 \\
x0.50 & \textbf{99.8} & 98.8 & 94.3 & \textbf{95.8} & \textbf{98.3} & 97.5 & \textbf{97.5} & 97.4 \\
x0.75 & \textbf{99.7} & 99.0 & 95.7 & \textbf{96.1} & \textbf{98.4} & 97.6 & \textbf{97.9} & 97.6 \\
\midrule
x1.00 & \textbf{99.3} & 99.0 & \textbf{97.0} & 96.2 & \textbf{98.2} & 97.6 & \textbf{98.2} & 97.6 \\
\bottomrule
\end{tabular}
\end{center}
\end{table}

\setlength{\tabcolsep}{2pt}
\begin{table}[t]
\begin{center}
\caption{Details of the variants of the Encoder. We report the number of each variant's parameters in millions (Params). ``Dim.'' denotes the corresponding dimension.}
\label{table:ab_variants}
\begin{tabular}{rccccc}
\toprule
Encoder & Layers & Hidden Dim. & Head Dim. & Heads & Params \\
\midrule
Small & 8 & 128 & 16 & 16 & 1.33 \\
Medium & 16 & 256 & 16 & 16 & 3.20 \\
Large & 16 & 512 & 16 & 16 & 8.55 \\
\bottomrule
\end{tabular}
\end{center}
\end{table}

\setlength{\tabcolsep}{2pt}
\begin{table}[t]
\begin{center}
\caption{Generalization evaluation of the proposed JCSAT. We train our model individually on the BUMocap, BUMocap-X and Shelf datasets and test it on others. The MPJPE (mm) is reported.}
\label{table:ab_gen}
\begin{tabular}{cccccc}
\toprule
\multicolumn{3}{c}{Training Dataset} & \multicolumn{3}{c}{MPJPE $\downarrow$} \\
BUMocap & BUMocap-X & Shelf & BUMocap & BUMocap-X & Shelf \\
\midrule
\cmark & & & 52.2 & 64.0 & 59.8 \\
& \cmark & & 66.0 & 49.7 & 61.3 \\
& & \cmark & 83.6 & 68.3 & 57.3 \\
\bottomrule
\end{tabular}
\end{center}
\end{table}

\subsection{Computational complexity analysis}

We reported the framework's inference speed \red{(the processing time per frame from 2D to 3D in milliseconds (\textit{ms}))} and compared it with other state-of-the-art methods in Table~\ref{table:speed}. \red{We also reported the time consumption of other steps: 1) 2D pose detection would take 5 \textit{ms}, 2) Joint Cloud construction would take 3 \textit{ms}.} Our framework was tested on an Intel i7-8700 3.20 GHz CPU and one NVIDIA RTX3090 GPU. Since some works didn't release the code, we can only directly reference the performance metrics as reported in the original papers. Table~\ref{table:speed} presents the inference speed of three optimization-based approaches~\cite{zhang20204d,dong2021fast,zhou2022quickpose} and five learning-based approaches~\cite{reddy2021tessetrack,lin2021multi,wu2021graph,jiang2022dual,ye2022faster} (Our framework falls into the second category) \red{as well as the performance of our model when the number of input views is reduced. It reveals that recent learning-based methods are generally slower than optimization-based methods. Our method demonstrates a rapid increase in inference speed as the number of input views decreases, underscoring its potential for acceleration.}

\red{\textbf{Remark.} We would like to clarify that, while QuickPose~\cite{zhou2022quickpose} appears to operate at a higher speed than our method with a small accuracy decrease of 0.1dB, such advantage in speed shall be attributed to professional programming optimization, such as the GNU C++ compiler~\cite{gnu}.  
However, learning-based approaches like ours typically utilize PyTorch~\cite{paszke2019pytorch} with costly floating-point computations. Therefore, a direct comparison of computational costs between these two method groups would be unfair. What's more important, since the Shelf dataset only contains relatively easy scenarios, the actual performance gap is not as trivial as Table~\ref{table:speed} suggests. This is what motivates us to build a fully annotated dataset, BUMocap-X, with heavy occlusions and complex interactions. 
Table~\ref{table:x_eval_2} showcases our framework's capability, which excels all compared ones by an obvious margin. It also shows that learning-based approaches (i.e., DMAE~\cite{jiang2022dual}, Faster VoxelPose~\cite{ye2022faster} and TEMPO~\cite{choudhury2023tempo}) are superior to optimization-based approaches (4DAssocGraph~\cite{zhang20204d}) under this more challenging dataset with heavy occlusions.}

\subsection{Ablation Study}

\noindent\textbf{Masked Learning and Kinematic Loss.}
We first investigated the impact of the Joint Cloud masked learning as well as the designed Kinematic Loss in rows 1 to 3 of Table.~\ref{table:ab_pcp} by PCP metrics on the Shelf dataset. We observed that when we remove the masking procedure, the PCP score drops 4\%. When the Kinematic Loss is removed, the PCP score drops 0.4\%. These results show that the Joint Cloud masking procedure increases the robustness of the model and the Kinematic Loss further improves its performance.

\noindent\textbf{Observation number.} As mentioned, Joint Cloud collects information from all estimated 2D joints. As our model reasons on trajectile, skeletal and angular information, we assume it gains robustness to ambiguities caused by occlusions. To validate our assumption, we decreased the observation number and \red{compared} the PCP scores on the Shelf dataset. Rows 4 to 6 of Table~\ref{table:ab_pcp} demonstrate that our model can overcome ambiguities caused by occlusions.

\noindent\textbf{Inaccurate camera parameters.} 
\red{We further assessed our model's resilience to inaccurate camera parameters and compared the results with 4DAssocGraph~\cite{zhang20204d}. We denote the view with inaccurate camera parameters as \textit{inaccurate view}. We intentionally add random errors to the camera parameters which causes random translation offsets within the range $\left [ -0.25cm, 0.25cm \right ]$ and random rotations around a random axis within the range $\left [ -2.5^{\circ}, 2.5^{\circ} \right ]$. Results in Table~\ref{table:ab_incorrect_camera} show that our model remains stable and outperforms 4DAssocGraph with no more than two \textit{inaccurate views}. Both of our method and 4DAssocGraph degrade seriously with three \textit{inaccurate views}. This noise causes the triangulation step to fail in the early stages of both pipelines.}

\noindent\textbf{Different input resolution.} 
\red{Considering that reducing the size of input images may result in the loss of semantic information and consequently degrade the performance of the 2D detector. We then compared our model against 4DAssocGraph~\cite{zhang20204d} with different rescaling factors. As shown in Table~\ref{table:ab_res}, ours exhibits superior robustness. Notably, when the rescaling factor is set to 0.5, our model achieves higher PCP scores for A1 and A3 than with a factor of 1.0. This suggests that our model has learnt motion prior, compensating for the missing information.}

\noindent\textbf{Feature aggregation.} 
\red{The impact of different feature aggregation methods in the proposed OTAP module is shown in rows 1 to 3 of} Table~\ref{table:ab_aggregate}. We investigated several aggregation combinations described in Sec.~\ref{sec:jcsa}: 
a) all OTAPs adopt non-selective aggregation (OTAP-a); 
b) $T_{\mathrm{OTAP}}$ and $S_{\mathrm{OTAP}}$ adopt non-selective aggregation and $V_{\mathrm{OTAP}}$ adopts selective aggregation (OTAP-b); 
c) $T_{\mathrm{OTAP}}$ and $S_{\mathrm{OTAP}}$ adopt selective aggregation and $V_\mathrm{OTAP}$ adopts non-selective aggregation (OTAP-c); 
d) all OTAPs adopt selective aggregation ($T_{\mathrm{Enc}}$-M\&$S_{\mathrm{Enc}}$-M, i.e. our final model). 
We observed that replacing the selective aggregation with non-selective aggregation methods leads to the most 2\% performance drop on PCP and 4.9\% in terms of MPJPE. The changes to Precision and Recall are not obvious. Results show the optimal transport framework helps JCSAT to select the most suitable candidate representation and further improves its performance.

\noindent\textbf{Variants of the Encoder.}
\red{We assessed our model over different configurations, as summarized in Table~\ref{table:ab_variants}. There are three variant designs for both the Trajectory Encoder $T_{\mathrm{Enc}}$ and the Structure Encoder $S_{\mathrm{Enc}}$. We reported the corresponding performance in rows 4 to 10 of Table~\ref{table:ab_aggregate}, where $T_{\mathrm{Enc}}$-S, $T_{\mathrm{Enc}}$-M and $T_{\mathrm{Enc}}$-L denote the Trajectory Encoder applied in ``Small'', ``Medium'' and ``Large'' configurations respectively. Other notations such as $S_{\mathrm{Enc}}$-S follow the same naming rules. Results in Table~\ref{table:ab_aggregate} reveal that $S_{\mathrm{Enc}}$ is more influential in the overall performance of the model. This is evidenced by the larger performance gains observed with increases in $S_{\mathrm{Enc}}$'s parameters compared to those in $T_{\mathrm{Enc}}$. The combination of $T_{\mathrm{Enc}}$-M and $S_{\mathrm{Enc}}$-M yields the most balanced outcomes in terms of computing complexity and accuracy.
}

\section{Limitations and Future directions}

As mentioned in Sec.~\ref{sec:enc}, Joint Cloud represents a 5-dimensional matrix, formulated as $\mathcal{J} \in \mathbb{R}^{N \times \dot{V} \times \dot{M} \times  I \times 3}$. During the process of encoding, each encoder operates on three dimensions. For instance, $T_{\mathrm{Enc}}$ takes $N \times \dot{M} \times dim$ as input. In practical implementation, we set $N=10$, $\dot{M}=4$ for one trajectory and there are 25 trajectories in total. Given the significant computational complexity associated with this setup, the proposed framework currently cannot run in \red{real-time}. 
Recent works~\cite{li2022exploiting,zhu2021posegtac} propose several encoding strategies to reduce the computing complexity, e.g., 1) use fewer tokens for representation, and 2) process data on the temporal and spatial domains in the same memory buffer. We will try to use fewer tokens to represent the motion, with each token denoting a trajectory or skeletal structure. Further acceleration can be achieved using TensorRT~\cite{tensorrt} and frame-skipping strategies.
To deal with even more serious occlusion, we will further investigate applying even more robust priors to better learn structures from noisy observations, such as fully articulated humanoid models (e.g., the SMPL-family~\cite{loper2015smpl,prokudin2021smplpix}).

\section{Conclusions}

In this paper, we have first introduced the concept of 
Joint Cloud which contains 3D candidates triangulated from 2D detections regardless of their ID in every view pair. This data arrangement facilitates information utilization from all observation angles.  
Since Joint Cloud is noisy and redundant, we have proposed a Joint Cloud Selection and Aggregation Transformer (JCSAT) framework to deeply explore the trajectile, skeletal and angular \red{correlations} among all 3D candidates derived from the Joint Cloud.
Different from existing Transformer-based frameworks that fully rely on self-attention mechanisms to aggregate features, we have proposed the Optimal Token Attention Path (OTAP) module implemented by a trainable optimal transport framework to select the most informative \red{features}. Furthermore, to validate the effectiveness of the proposed framework, we have built and published a new multi-person motion capture dataset \red{featuring} complex interactions and severe occlusions, named BUMocap-X. 
Extensive experiments on public datasets as well as our BUMocap-X dataset establish the superiority of our approach over state-of-the-art methods.

\ifCLASSOPTIONcaptionsoff
  \newpage
\fi




%

\bibliographystyle{IEEEtran}
\normalem
\bibliography{mybib}

\begin{thebibliography}{10}
\providecommand{\url}[1]{#1}
\csname url@samestyle\endcsname
\providecommand{\newblock}{\relax}
\providecommand{\bibinfo}[2]{#2}
\providecommand{\BIBentrySTDinterwordspacing}{\spaceskip=0pt\relax}
\providecommand{\BIBentryALTinterwordstretchfactor}{4}
\providecommand{\BIBentryALTinterwordspacing}{\spaceskip=\fontdimen2\font plus
\BIBentryALTinterwordstretchfactor\fontdimen3\font minus
  \fontdimen4\font\relax}
\providecommand{\BIBforeignlanguage}[2]{{%
\expandafter\ifx\csname l@#1\endcsname\relax
\typeout{** WARNING: IEEEtran.bst: No hyphenation pattern has been}%
\typeout{** loaded for the language `#1'. Using the pattern for}%
\typeout{** the default language instead.}%
\else
\language=\csname l@#1\endcsname
\fi
#2}}
\providecommand{\BIBdecl}{\relax}
\BIBdecl

\bibitem{vive}
VIVE, ``Vive | discover virtual reality beyond imagination,''
  \url{https://www.vive.com/}, 2022, accessed December 29, 2022.

\bibitem{hololens}
Microsoft, ``Microsoft hololens | mixed reality technology for business,''
  \url{https://www.microsoft.com/en-us/hololens/hardware}, 2022, accessed
  December 29, 2022.

\bibitem{oculus}
Meta, ``Meta quest vr headsets, accessories \& equipment,''
  \url{https://www.meta.com/quest/}, 2022, accessed January 31, 2023.

\bibitem{optitrack}
OptiTrack, ``Optitrack - motion capture systems,''
  \url{https://optitrack.com/}, 2019, accessed August 19, 2021.

\bibitem{vicon}
Vicon, ``Vicon | award winning motion capture systems,''
  \url{https://www.vicon.com/}, 2022, accessed December 29, 2022.

\bibitem{xsens}
Xsens, ``Xsens 3d motion tracking,'' \url{https://www.xsens.com/}, 2022,
  accessed December 29, 2022.

\bibitem{cao2019openpose}
Z.~Cao, G.~Hidalgo, T.~Simon, S.-E. Wei, and Y.~Sheikh, ``{OpenPose}: realtime
  multi-person 2d pose estimation using part affinity fields,'' \emph{IEEE
  Transactions on Pattern Analysis and Machine Intelligence}, vol.~43, no.~1,
  pp. 172--186, 2019.

\bibitem{wu2019detectron2}
Y.~Wu, A.~Kirillov, F.~Massa, W.-Y. Lo, and R.~Girshick, ``Detectron2,''
  \url{https://github.com/facebookresearch/detectron2}, 2019.

\bibitem{sun2019deep}
K.~Sun, B.~Xiao, D.~Liu, and J.~Wang, ``Deep high-resolution representation
  learning for human pose estimation,'' in \emph{Proceedings of the IEEE/CVF
  conference on computer vision and pattern recognition}, 2019, pp. 5693--5703.

\bibitem{fang2022alphapose}
H.-S. Fang, J.~Li, H.~Tang, C.~Xu, H.~Zhu, Y.~Xiu, Y.-L. Li, and C.~Lu,
  ``Alphapose: Whole-body regional multi-person pose estimation and tracking in
  real-time,'' \emph{IEEE Transactions on Pattern Analysis and Machine
  Intelligence}, pp. 7157--7173, 2022.

\bibitem{pavllo20193d}
D.~Pavllo, C.~Feichtenhofer, D.~Grangier, and M.~Auli, ``3d human pose
  estimation in video with temporal convolutions and semi-supervised
  training,'' in \emph{Proceedings of the IEEE/CVF Conference on Computer
  Vision and Pattern Recognition}, 2019, pp. 7753--7762.

\bibitem{kolotouros2019learning}
N.~Kolotouros, G.~Pavlakos, M.~J. Black, and K.~Daniilidis, ``Learning to
  reconstruct 3d human pose and shape via model-fitting in the loop,'' in
  \emph{Proceedings of the IEEE/CVF International Conference on Computer
  Vision}, 2019, pp. 2252--2261.

\bibitem{jiang2024explore}
J.~Jiang and J.~Chen, ``Exploring latent cross-channel embedding for accurate
  3d human pose reconstruction in a diffusion framework,'' in \emph{2024 IEEE
  International Conference on Acoustics, Speech and Signal Processing}, 2024,
  pp. 7870--7874.

\bibitem{zhang20204d}
Y.~Zhang, L.~An, T.~Yu, X.~Li, K.~Li, and Y.~Liu, ``{4D} association graph for
  realtime multi-person motion capture using multiple video cameras,'' in
  \emph{IEEE/CVF Conference on Computer Vision and Pattern Recognition}, 2020,
  pp. 1324--1333.

\bibitem{dong2021fast}
J.~Dong, Q.~Fang, W.~Jiang, Y.~Yang, Q.~Huang, H.~Bao, and X.~Zhou, ``Fast and
  robust multi-person 3d pose estimation and tracking from multiple views,''
  \emph{IEEE Transactions on Pattern Analysis and Machine Intelligence},
  vol.~14, no.~8, pp. 1--1, 2021.

\bibitem{tu2020voxelpose}
H.~Tu, C.~Wang, and W.~Zeng, ``Voxelpose: Towards multi-camera 3d human pose
  estimation in wild environment,'' in \emph{European Conference on Computer
  Vision}, 2020, pp. 197--212.

\bibitem{li2022exploiting}
W.~Li, H.~Liu, R.~Ding, M.~Liu, P.~Wang, and W.~Yang, ``Exploiting temporal
  contexts with strided transformer for 3d human pose estimation,'' \emph{IEEE
  Transactions on Multimedia}, vol.~3, no.~1, pp. 1--1, 2022.

\bibitem{zhou2022quickpose}
Z.~Zhou, Q.~Shuai, Y.~Wang, Q.~Fang, X.~Ji, F.~Li, H.~Bao, and X.~Zhou,
  ``Quickpose: Real-time multi-view multi-person pose estimation in crowded
  scenes,'' in \emph{In proceedings of SIGGRAPH}, 2022, pp. 1--9.

\bibitem{hartley2003multiple}
R.~Hartley and A.~Zisserman, \emph{Multiple view geometry in computer
  vision}.\hskip 1em plus 0.5em minus 0.4em\relax Cambridge University Press,
  2003.

\bibitem{belagiannis2014multiple}
V.~Belagiannis, X.~Wang, B.~Schiele, P.~Fua, S.~Ilic, and N.~Navab, ``Multiple
  human pose estimation with temporally consistent 3d pictorial structures,''
  in \emph{European Conference on Computer Vision}, 2014, pp. 742--754.

\bibitem{reddy2021tessetrack}
N.~D. Reddy, L.~Guigues, L.~Pishchulin, J.~Eledath, and S.~G. Narasimhan,
  ``Tessetrack: End-to-end learnable multi-person articulated 3d pose
  tracking,'' in \emph{Proceedings of the IEEE/CVF Conference on Computer
  Vision and Pattern Recognition}, 2021, pp. 15\,190--15\,200.

\bibitem{wu2021graph}
S.~Wu, S.~Jin, W.~Liu, L.~Bai, C.~Qian, D.~Liu, and W.~Ouyang, ``Graph-based 3d
  multi-person pose estimation using multi-view images,'' in \emph{Proceedings
  of the IEEE/CVF International Conference on Computer Vision}, 2021, pp.
  11\,148--11\,157.

\bibitem{jiang2022dual}
J.~Jiang, J.~Chen, and Y.~Guo, ``A dual-masked auto-encoder for robust motion
  capture with spatial-temporal skeletal token completion,'' in
  \emph{Proceedings of the ACM International Conference on Multimedia}, 2022,
  pp. 5123--5131.

\bibitem{he2022masked}
K.~He, X.~Chen, S.~Xie, Y.~Li, P.~Doll{\'a}r, and R.~Girshick, ``Masked
  autoencoders are scalable vision learners,'' in \emph{Proceedings of the
  IEEE/CVF Conference on Computer Vision and Pattern Recognition}, 2022, pp.
  16\,000--16\,009.

\bibitem{choudhury2023tempo}
R.~Choudhury, K.~M. Kitani, and L.~A. Jeni, ``Tempo: Efficient multi-view pose
  estimation, tracking, and forecasting,'' in \emph{Proceedings of the IEEE/CVF
  International Conference on Computer Vision}, 2023, pp. 14\,750--14\,760.

\bibitem{devlin2018bert}
J.~Devlin, M.-W. Chang, K.~Lee, and K.~Toutanova, ``Bert: Pre-training of deep
  bidirectional transformers for language understanding,'' \emph{arXiv preprint
  arXiv:1810.04805}, 2018.

\bibitem{belagiannis20143d}
V.~Belagiannis, S.~Amin, M.~Andriluka, B.~Schiele, N.~Navab, and S.~Ilic, ``3d
  pictorial structures for multiple human pose estimation,'' in
  \emph{Proceedings of the IEEE Conference on Computer Vision and Pattern
  Recognition}, 2014, pp. 1669--1676.

\bibitem{ye2022faster}
H.~Ye, W.~Zhu, C.~Wang, R.~Wu, and Y.~Wang, ``Faster voxelpose: Real-time 3d
  human pose estimation by orthographic projection,'' in \emph{European
  Conference on Computer Vision}.\hskip 1em plus 0.5em minus 0.4em\relax
  Springer, 2022, pp. 142--159.

\bibitem{srivastav2024selfpose3d}
V.~Srivastav, K.~Chen, and N.~Padoy, ``Selfpose3d: Self-supervised multi-person
  multi-view 3d pose estimation,'' \emph{arXiv preprint arXiv:2404.02041},
  2024.

\bibitem{lin2021multi}
J.~Lin and G.~H. Lee, ``Multi-view multi-person 3d pose estimation with plane
  sweep stereo,'' in \emph{Proceedings of the IEEE/CVF Conference on Computer
  Vision and Pattern Recognition}, 2021, pp. 11\,886--11\,895.

\bibitem{rempe2021humor}
D.~Rempe, T.~Birdal, A.~Hertzmann, J.~Yang, S.~Sridhar, and L.~J. Guibas,
  ``Humor: 3d human motion model for robust pose estimation,'' in
  \emph{Proceedings of the IEEE/CVF International Conference on Computer
  Vision}, 2021, pp. 11\,488--11\,499.

\bibitem{huang2022neural}
B.~Huang, L.~Pan, Y.~Yang, J.~Ju, and Y.~Wang, ``Neural mocon: Neural motion
  control for physically plausible human motion capture,'' in \emph{Proceedings
  of the IEEE/CVF Conference on Computer Vision and Pattern Recognition}, 2022,
  pp. 6417--6426.

\bibitem{gong2022posetriplet}
K.~Gong, B.~Li, J.~Zhang, T.~Wang, J.~Huang, M.~B. Mi, J.~Feng, and X.~Wang,
  ``Posetriplet: Co-evolving 3d human pose estimation, imitation, and
  hallucination under self-supervision,'' in \emph{Proceedings of the IEEE/CVF
  Conference on Computer Vision and Pattern Recognition}, 2022, pp.
  11\,017--11\,027.

\bibitem{maeda2022motionaug}
T.~Maeda and N.~Ukita, ``Motionaug: Augmentation with physical correction for
  human motion prediction,'' in \emph{Proceedings of the IEEE/CVF Conference on
  Computer Vision and Pattern Recognition}, 2022, pp. 6427--6436.

\bibitem{wandt2022elepose}
B.~Wandt, J.~J. Little, and H.~Rhodin, ``Elepose: Unsupervised 3d human pose
  estimation by predicting camera elevation and learning normalizing flows on
  2d poses,'' in \emph{Proceedings of the IEEE/CVF Conference on Computer
  Vision and Pattern Recognition}, 2022, pp. 6635--6645.

\bibitem{wandt2021canonpose}
B.~Wandt, M.~Rudolph, P.~Zell, H.~Rhodin, and B.~Rosenhahn, ``Canonpose:
  Self-supervised monocular 3d human pose estimation in the wild,'' in
  \emph{Proceedings of the IEEE/CVF Conference on Computer Vision and Pattern
  Recognition}, 2021, pp. 13\,294--13\,304.

\bibitem{huang2022pose2uv}
B.~Huang, T.~Zhang, and Y.~Wang, ``Pose2uv: Single-shot multiperson mesh
  recovery with deep uv prior,'' \emph{IEEE Transactions on Image Processing},
  vol.~31, pp. 4679--4692, 2022.

\bibitem{loper2015smpl}
M.~Loper, N.~Mahmood, J.~Romero, G.~Pons-Moll, and M.~J. Black, ``{SMPL}: A
  skinned multi-person linear model,'' \emph{ACM Transactions on graphics},
  vol.~34, no.~6, pp. 1--16, 2015.

\bibitem{yan2018spatial}
S.~Yan, Y.~Xiong, and D.~Lin, ``Spatial temporal graph convolutional networks
  for skeleton-based action recognition,'' \emph{Proceedings of the AAAI
  Conference on Artificial Intelligence}, vol.~32, no.~1, pp. 3482--3489, 2018.

\bibitem{zheng20213d}
C.~Zheng, S.~Zhu, M.~Mendieta, T.~Yang, C.~Chen, and Z.~Ding, ``3d human pose
  estimation with spatial and temporal transformers,'' in \emph{Proceedings of
  the IEEE/CVF International Conference on Computer Vision}, 2021, pp.
  11\,656--11\,665.

\bibitem{zhu2021posegtac}
Y.~Zhu, X.~Xu, F.~Shen, Y.~Ji, L.~Gao, and H.~T. Shen, ``Posegtac: Graph
  transformer encoder-decoder with atrous convolution for 3d human pose
  estimation.'' in \emph{Proceedings of the International Joint Conference on
  Artificial Intelligence}, 2021, pp. 1359--1365.

\bibitem{vaswani2017attention}
A.~Vaswani, N.~Shazeer, N.~Parmar, J.~Uszkoreit, L.~Jones, A.~N. Gomez, L.~u.
  Kaiser, and I.~Polosukhin, ``Attention is all you need,'' in \emph{Advances
  in Neural Information Processing Systems}, vol.~30, 2017, pp. 1--1.

\bibitem{park2023towards}
S.~Park, E.~You, I.~Lee, and J.~Lee, ``Towards robust and smooth 3d
  multi-person pose estimation from monocular videos in the wild,'' in
  \emph{Proceedings of the IEEE/CVF International Conference on Computer
  Vision}, 2023, pp. 14\,772--14\,782.

\bibitem{shuai2022adaptive}
H.~Shuai, L.~Wu, and Q.~Liu, ``Adaptive multi-view and temporal fusing
  transformer for 3d human pose estimation,'' \emph{IEEE Transactions on
  Pattern Analysis and Machine Intelligence}, vol.~45, no.~4, pp. 4122--4135,
  2022.

\bibitem{tancik2020fourier}
M.~Tancik, P.~Srinivasan, B.~Mildenhall, S.~Fridovich-Keil, N.~Raghavan,
  U.~Singhal, R.~Ramamoorthi, J.~Barron, and R.~Ng, ``Fourier features let
  networks learn high frequency functions in low dimensional domains,''
  \emph{Advances in Neural Information Processing Systems}, vol.~33, pp.
  7537--7547, 2020.

\bibitem{dosovitskiy2020image}
A.~Dosovitskiy, L.~Beyer, A.~Kolesnikov, D.~Weissenborn, X.~Zhai,
  T.~Unterthiner, M.~Dehghani, M.~Minderer, G.~Heigold, S.~Gelly, J.~Uszkoreit,
  and N.~Houlsby, ``An image is worth 16x16 words: Transformers for image
  recognition at scale,'' in \emph{International Conference on Learning
  Representations}, 2021, pp. 1--1.

\bibitem{mialon2020trainable}
G.~Mialon, D.~Chen, A.~d’Aspremont, and J.~Mairal, ``A trainable optimal
  transport embedding for feature aggregation,'' in \emph{International
  Conference on Learning Representations}, 2020, pp. 1--1.

\bibitem{sinkhorn1967diagonal}
R.~Sinkhorn, ``Diagonal equivalence to matrices with prescribed row and column
  sums,'' \emph{The American Mathematical Monthly}, vol.~74, no.~4, pp.
  402--405, 1967.

\bibitem{joo2015panoptic}
H.~Joo, H.~Liu, L.~Tan, L.~Gui, B.~Nabbe, I.~Matthews, T.~Kanade, S.~Nobuhara,
  and Y.~Sheikh, ``Panoptic studio: A massively multiview system for social
  motion capture,'' in \emph{Proceedings of the IEEE International Conference
  on Computer Vision}, 2015, pp. 3334--3342.

\bibitem{azurekinect}
Microsoft, ``Azure kinect dk - build for mixed reality using ai sensors,''
  \url{https://azure.microsoft.com/en-us/services/kinect-dk/}, 2019, accessed
  January 29, 2022.

\bibitem{belagiannis20153d}
V.~Belagiannis, S.~Amin, M.~Andriluka, B.~Schiele, N.~Navab, and S.~Ilic,
  ``{3D} pictorial structures revisited: Multiple human pose estimation,''
  \emph{IEEE Transactions on Pattern Analysis and Machine Intelligence},
  vol.~38, no.~10, pp. 1929--1942, 2015.

\bibitem{lin2014microsoft}
T.-Y. Lin, M.~Maire, S.~Belongie, J.~Hays, P.~Perona, D.~Ramanan,
  P.~Doll{\'a}r, and C.~L. Zitnick, ``Microsoft coco: Common objects in
  context,'' in \emph{European conference on computer vision}, 2014, pp.
  740--755.

\bibitem{gnu}
Wikipedia, ``Gnu compiler collection,''
  \url{https://en.wikipedia.org/wiki/GNU_Compiler_Collection/}, 2024, accessed
  May 29, 2024.

\bibitem{paszke2019pytorch}
A.~Paszke, S.~Gross, F.~Massa, A.~Lerer, J.~Bradbury, G.~Chanan, T.~Killeen,
  Z.~Lin, N.~Gimelshein, L.~Antiga \emph{et~al.}, ``Pytorch: An imperative
  style, high-performance deep learning library,'' \emph{Advances in neural
  information processing systems}, vol.~32, 2019.

\bibitem{tensorrt}
Nvidia, ``Tensorrt sdk - nvidia developer,''
  \url{https://developer.nvidia.com/tensorrt}, 2022, accessed June 11, 2023.

\bibitem{prokudin2021smplpix}
S.~Prokudin, M.~J. Black, and J.~Romero, ``{SMPLpix}: Neural avatars from 3d
  human models,'' in \emph{IEEE Winter Conference on Applications of Computer
  Vision}, 2021, pp. 1810--1819.

\bibitem{pavlakos2019expressive}
G.~Pavlakos, V.~Choutas, N.~Ghorbani, T.~Bolkart, A.~A. Osman, D.~Tzionas, and
  M.~J. Black, ``Expressive body capture: {3D} hands, face, and body from a
  single image,'' in \emph{IEEE/CVF Conference on Computer Vision and Pattern
  Recognition}, 2019, pp. 10\,975--10\,985.

\bibitem{goyal2017accurate}
P.~Goyal, P.~Doll{\'a}r, R.~Girshick, P.~Noordhuis, L.~Wesolowski, A.~Kyrola,
  A.~Tulloch, Y.~Jia, and K.~He, ``Accurate, large minibatch sgd: Training
  imagenet in 1 hour,'' \emph{arXiv preprint arXiv:1706.02677}, 2017.

\bibitem{loshchilov2017decoupled}
I.~Loshchilov and F.~Hutter, ``Decoupled weight decay regularization,''
  \emph{arXiv preprint arXiv:1711.05101}, 2017.

\bibitem{loshchilov2016sgdr}
------, ``Sgdr: Stochastic gradient descent with warm restarts,'' \emph{arXiv
  preprint arXiv:1608.03983}, 2016.

\end{thebibliography}


%





\begin{IEEEbiography}[{\includegraphics[width=1in,height=1.25in,clip,keepaspectratio]{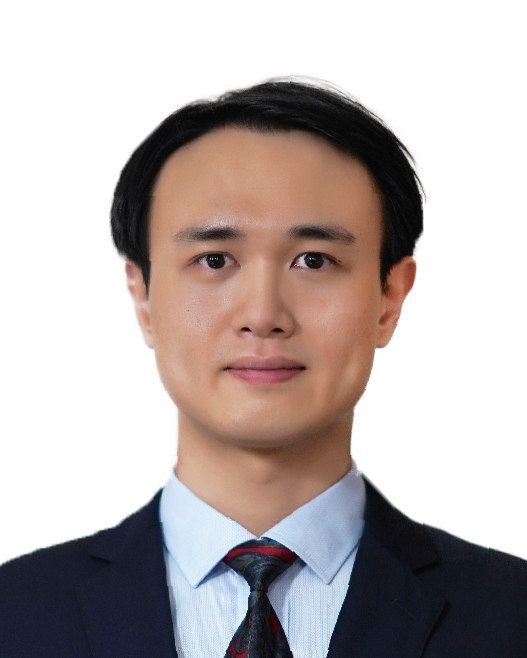}}]{Junkun Jiang} is currently a Ph.D. student at the Department of Computer Science, Hong Kong Baptist University. He received the B.Eng. degree in Computer Science and Technology from Xiamen University, China, and the M.Eng. degree in Software Engineering from Sun Yat-sen University, China, in 2017 and 2019 respectively. He worked as a senior algorithm engineer at Bigo Technology Pte. Ltd., Guangzhou, China. His research interests are 3D motion capture, human motion estimation and analysis, and action recognition.
\end{IEEEbiography} \vfill

\begin{IEEEbiography}[{\includegraphics[width=1in,height=1.25in,clip,keepaspectratio]{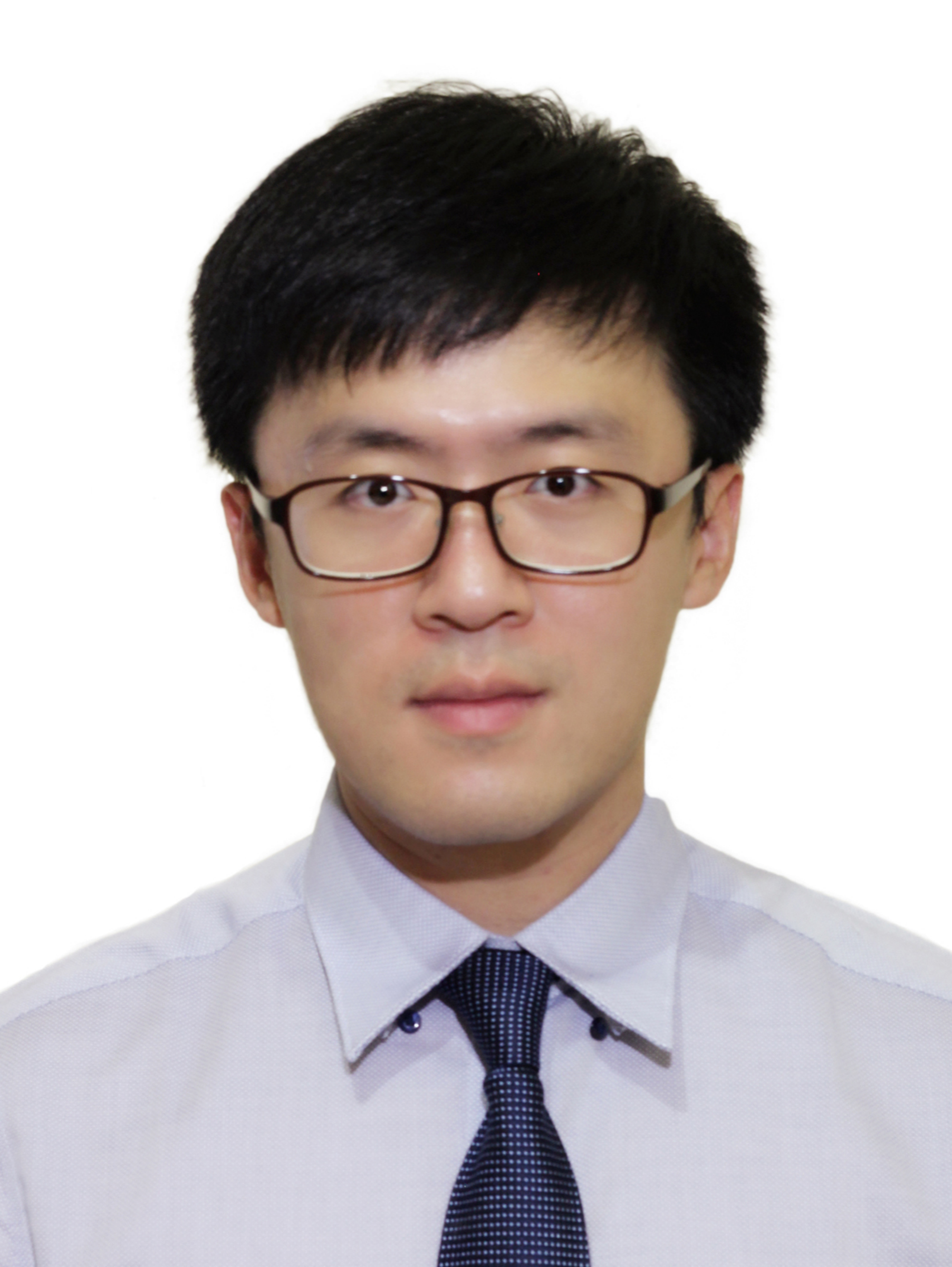}}]{Jie Chen} (Senior Member, IEEE) is currently an Assistant Professor at the Department of Computer Science, Hong Kong Baptist University. He received the B.Sc. and M. Eng. degrees both from the School of Optical and Electronic Information, Huazhong University of Science and Technology, China, and the Ph.D. degree from the School of Electrical and Electronic Engineering, Nanyang Technological University, Singapore in 2016. He worked as a post-doctoral research fellow at ST Engineering-NTU Corporate Laboratory, Singapore, and then as a senior algorithm engineer at OmniVision Technologies Inc. His research focuses on computational photography (light fields, high dynamic range imaging, hyperspectral imaging and computational tomography), multimedia signal capture, reconstruction and content generation (3D vision, motion and music), AI for Art-Tech and Humanities. He currently serves as an Associate Editor for the Visual Computer Journal, Springer.
\end{IEEEbiography} \vfill

\begin{IEEEbiography}[{\includegraphics[width=1in,height=1.25in,clip,keepaspectratio]{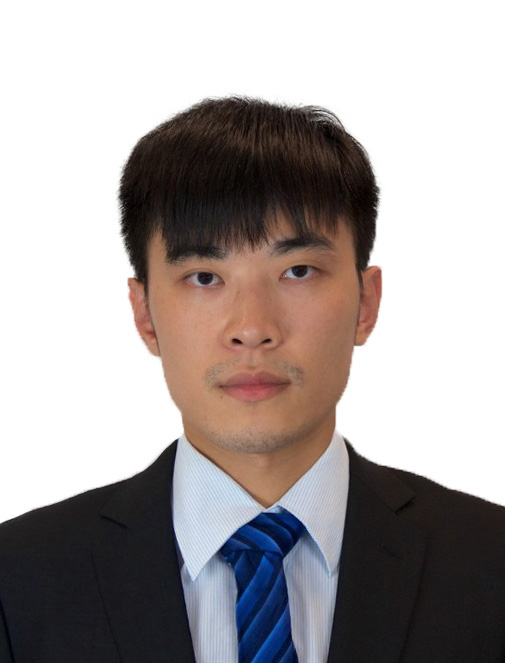}}]{Ho Yin Au} is currently a Ph.D. student at the Department of Computer Science, Hong Kong Baptist University, Hong Kong. He received the B.S. in Mathematics and Computer Science from University of Illinois Urbana-Champaign, USA, and M.S. in Computer Science from University of Massachusetts Amherst, USA, in 2018 and 2020 respectively. His research interests are human motion capture and modelling, multimedia cross-modal analysis, synthesis, and translation. He is a recipient of Hong Kong PhD Fellowship.
\end{IEEEbiography} \vfill

\begin{IEEEbiography}[{\includegraphics[width=1in,height=1.25in,clip,keepaspectratio]{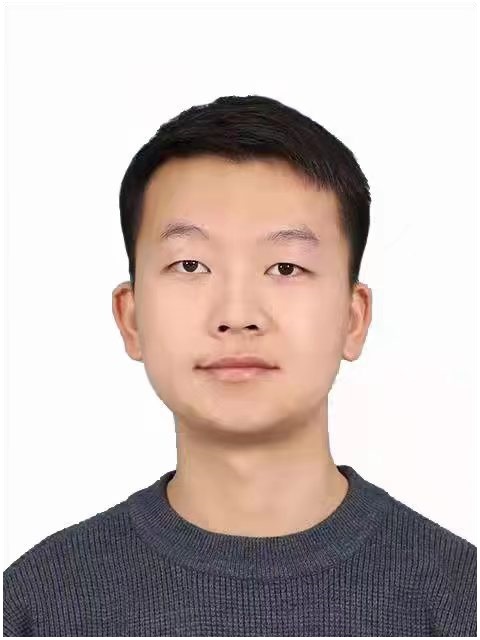}}]{Mingyuan Chen} is currently a research assistant at the Department of Computer Science, Hong Kong Baptist University. He received the B.Sc. degree in Computer Science from Tianjin Normal University, China, and the M.Sc. degree in Computer Science from Hong Kong Baptist University, China, in 2022 and 2023, respectively. His research interests are human motion estimation and generation.
\end{IEEEbiography} \vfill

\begin{IEEEbiography}[{\includegraphics[width=1in,height=1.25in,clip,keepaspectratio]{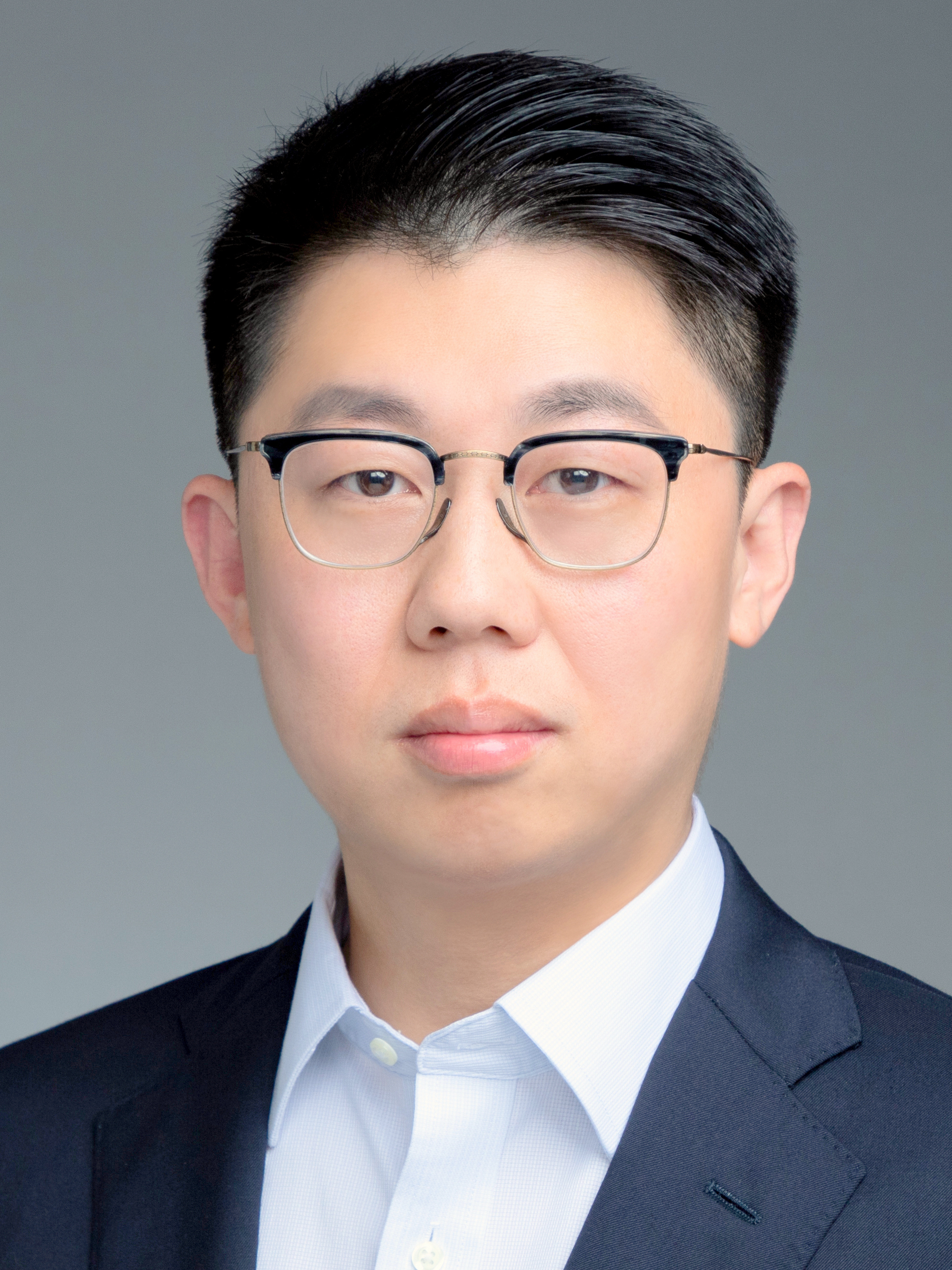}}]{Wei Xue} (Member, IEEE) is currently an Assistant Professor at Division of Emerging Interdisciplinary Areas (EMIA), Hong Kong University of Science and Technology (HKUST). He received the Bachelor degree in automatic control from Huazhong University of Science and Technology in 2010, and the Ph.D degree in pattern recognition and intelligent systems from Institute of Automation, Chinese Academy of Sciences in 2015. From August 2015 to September 2018 he was first a Marie Curie Experienced Researcher and then a Research Associate in Speech and Audio Processing Group, Department of Electrical\&Electronic Engineering, Imperial College London, UK. He was a Senior Research Scientist at JD AI Research, Beijing, from November 2018 to December 2021, where he was leading the R\&D on front-end speech processing and acoustic modelling for robust speech recognition. From January 2022 to April 2023 he was an Assistant Professor at Department of Computer Sciences, Hong Kong Baptist University. He was a visiting scholar at Université de Toulon and KU Leuven. Wei's research interests are in speech and music intelligence, including AI music generation, speech enhancement and separation, room acoustics, as well as speech and audio event recognition. He was a former Marie Curie Fellow and was selected into the Beijing Overseas Talent Aggregation Project. 
\end{IEEEbiography} \vfill

\begin{IEEEbiography}[{\includegraphics[width=1in,height=1.25in,clip,keepaspectratio]{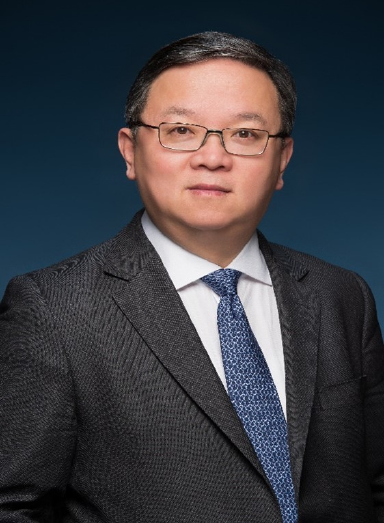}}]{Yike Guo} (Fellow, IEEE) is currently a Chair Professor in the Department of Computer Science and Engineering, the Hong Kong  University of Science and Technology, where he serves as the Provost since December 2022. He received his first-class honours degree in Computing Science from Tsinghua University in 1985 and obtained his PhD degree from Imperial College London in 1994. He has been a full Professor in the Department of Computing of Imperial College London since 2002. He was the Founding Director of the Data Science Institute at Imperial College London since 2014. In 2015 - 2020, Professor Guo was appointed as Non-Executive Dean of the School of Computer Engineering and Science in Shanghai University and he is now the Honorary Dean of the School. Prior to joining HKUST, Professor Guo was the Vice President (Research and Development) and the Dean of Graduate School at Hong Kong Baptist University since 2020.

Professor Guo's research focuses on machine learning and data mining for large-scale scientific applications including distributed data mining methods, machine learning and informatics systems for biology, chemistry, geophysics, healthcare, environment, economy, finance, social media, creative design and art applications. He has extensive experience in working with industries. He has led the development of start-up companies and worked with leading international companies such as GSK, Pfizer, Roche, KPMG, Huawei and BBC et al in large research projects and consulting services.

Professor Guo is Fellow of Royal Academy of Engineering (FREng), a Member of Academia Europaea (MAE), Fellow of Hong Kong Academy of Engineering Sciences (FHKEng), Fellow of the Institute of Electrical and Electronics Engineers (FIEEE), Fellow of British Computer Society (FBCS), and Fellow of Chinese Association for Artificial Intelligence (FCAAI). He has served on the editorial board of many first-tier journals. He is the editor-in-chief of Annual Reviews of Data Sciences (World Scientific), and the deputy editor-in-chief of CAAI Transections on Intelligent Systems (the official journal of CAAI), Machine Intelligence Research (Springer).
\end{IEEEbiography} \vfill

\clearpage

\twocolumn[
    \begin{@twocolumnfalse}
        \customtitle
    \end{@twocolumnfalse}
]
\setcounter{section}{0}
\setcounter{figure}{0}
\setcounter{table}{0}
\section{Joint Cloud construction example}

Here we give an example of the detailed construction of the proposed Joint Cloud. Images used here are from the Shelf dataset~\cite{pavlakos2019expressive}. We will explain the process step by step with visual demonstrations. For a more illustrative explanation, we provide a supplementary video containing joint cloud construction animation and visual comparison to other methods. 

\begin{figure}[t]
\centering
\includegraphics[width=0.98\linewidth]{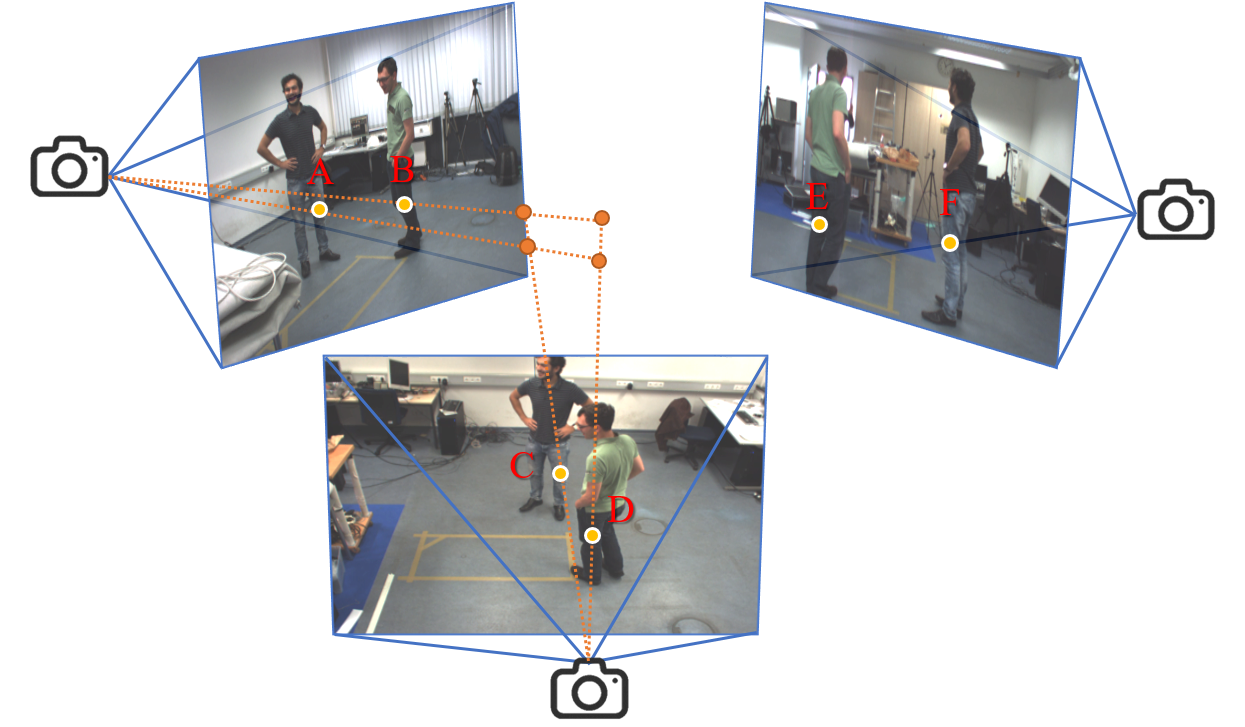}
\caption{Demonstration of the triangulation procedure in the Joint Cloud construction pipeline. Three views are presented in the scene. The 2D detections of the "Left Knee" are depicted by yellow dots. From one of the view pair combinations, the 3D triangulated candidates are shown in orange dots. Orange dashed lines symbolize the ray cast from a camera centre to the 2D point on the camera plane. For clarity, only four candidates are illustrated while there should be twelve candidates in total.}
\label{fig:jc_triangulate}
\end{figure}

\begin{figure}[t]
\centering
\includegraphics[width=0.98\linewidth]{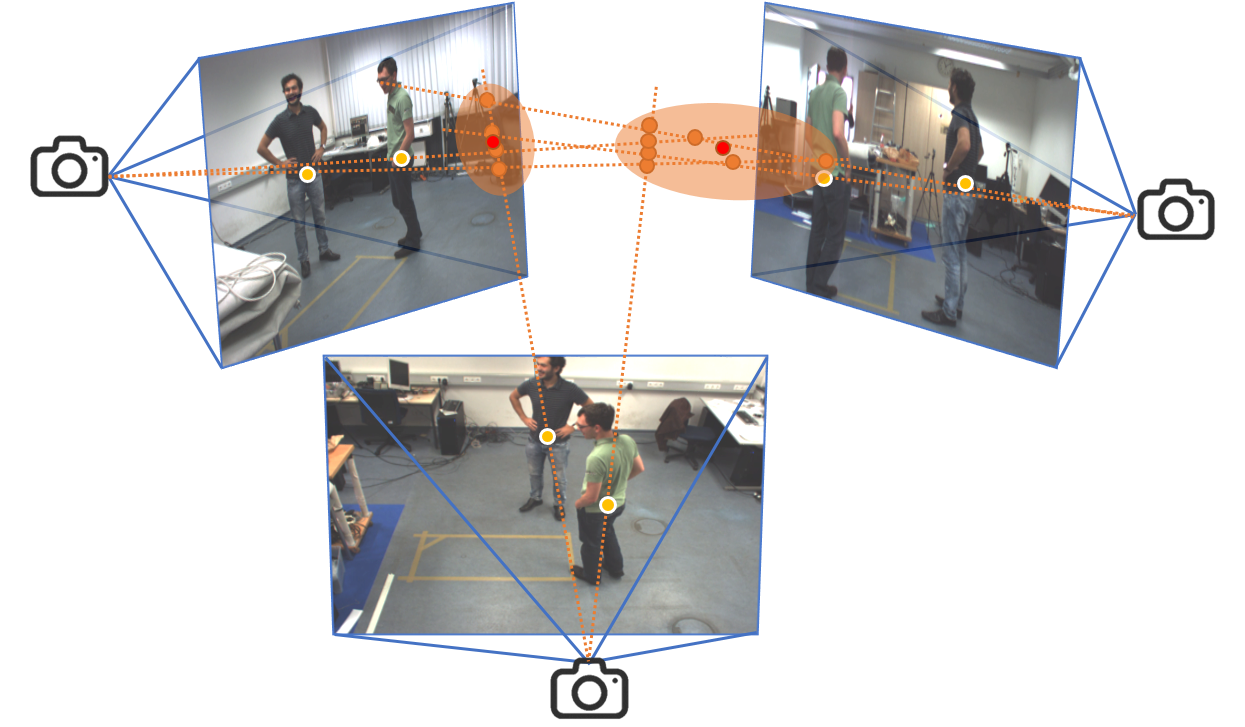}
\caption{Demonstration of the mid-hip clustering procedure in the Joint Cloud construction pipeline. Yellow dots represent the 2D detections of ``mid-hip''. Orange dots represent the 3D triangulated candidates. Red dots represent the centroid of each cluster. We utilize the centroid as each individual's body centre indicating the corresponding Joint Cloud subset's centre.}
\label{fig:jc_cluster}
\end{figure}

\subsection{Triangulation in view pairs}

As mentioned in the main paper, the Joint Cloud consists of noisy and redundant 3D joint candidates $j_{n,\dot{\mathbf{v}}}^{i,\dot{\mathbf{m}}} \in \mathbb{R}^3$, where subscript $n\in[1, N]$ denotes the frame number and $\dot{\mathbf{v}}\in[1, \dot{V}]$ represents one of the view pair combinations (Eq. 1 in the main paper), superscript $i\in[1, I]$ denotes the joint type and $\dot{\mathbf{m}}\in[1, \dot{M}]$ denotes joint's initial identity (Eq. 2 in the main paper) respectively. As such, 
the Joint Cloud can be formulated as:
\begin{align*}
\label{eq.id_comb}
\begin{split}
\mathbf{J}_n=\left\{j_{n,\dot{\mathbf{v}}}^{i,\dot{\mathbf{m}}} \in \mathbb{R}^3 \right\}, \\
\mathcal{J} = \left\{\mathbf{J}_1, \mathbf{J}_2, \cdots, \mathbf{J}_N\right\},
\end{split}
\end{align*}
where $\mathbf{J}_n$ stands for $n$-th frame's 3D candidate set and $\mathcal{J}$ is the Joint Cloud accumulated from $\mathbf{J}_n$. 

Fig.~\ref{fig:jc_triangulate} illustrates a scene with three views and one frame. If OpenPose~\cite{cao2019openpose} is used as the 2D pose detection module, the total number of joint type $I=25$ due to the 25-joint skeleton output format of the OpenPose~\cite{cao2019openpose}. As the number of the observed individuals is 2 in each view, every two views, a.k.a. one view pair combination observes $\dot{\mathbf{M}} = 2 \times 2 = 4$ ID combinations. $N=1$ cause there is only one frame. $\dot{\mathbf{V}}=3$ because there are three views and the view pair combination is $\begin{pmatrix} 3 \\ 2 \end{pmatrix}=3$. Let's focus on the joint type ``Left Knee''. Regardless of the individual's ID, literally, we have 6 ``Left Knee'' joints denoted as A, B, C, D, E, and F in three views. We can triangulate four candidates of ``Left Knee'' from each view pair combination $\dot{\mathbf{v}}$, e.g. A and C, A and D, B and C, B and D. Combining three view-pair combinations, the total number of candidates for ``Left Knee'' leads to 12. In Fig.~\ref{fig:jc_triangulate}, we only demonstrate one view combination's candidates for a clear view. Following the above instruction, we can triangulate tons of joint candidates from every frame in a video sequence. In the next subsection, we propose a simple clustering algorithm to assign initial IDs to joint candidates so that we get a smaller Joint Cloud.

\subsection{Assign initial IDs to Joint Cloud}

Since joint candidates are noisy and redundant in the Joint Cloud, we can adopt a rough filter to assign the initial target identity to those candidates thereby reducing the computational complexity. To illustrate, we know that a human leg cannot exceed a length of two meters, so we can filter out any limbs that surpass this limit. Following this intuition, we propose a simple but efficient clustering algorithm, outlined in Algorithm 1 of the main paper, to enable effective data reduction.
 
\subsubsection{Mid-hip clustering}\label{sec:cluster}

We need to verify the number of individuals in every frame so that we can divide the Joint Cloud for each individual. To achieve this, we employ the KMeans clustering algorithm, utilizing joint candidates of the ``mid-hip" type, to obtain a distinct central location for each individual. Initially, we establish the minimum number of individuals across views. Let's take the scene depicted in Fig.~\ref{fig:jc_cluster} as an example, the minimum individual number of this scene is 2. As mentioned in the main paper, the minimum individual number in one scene is the minimum number of detected persons across different views. Subsequently, we set the cluster number with the minimum individual number and cluster the mid-hip candidates. Lastly, after clustering, we assign the centroid of each cluster as each individual's central location. Therefore, we can divide those joint candidates with two IDs and we process them separately. 

\begin{figure}
\centering
\includegraphics[width=0.98\linewidth]{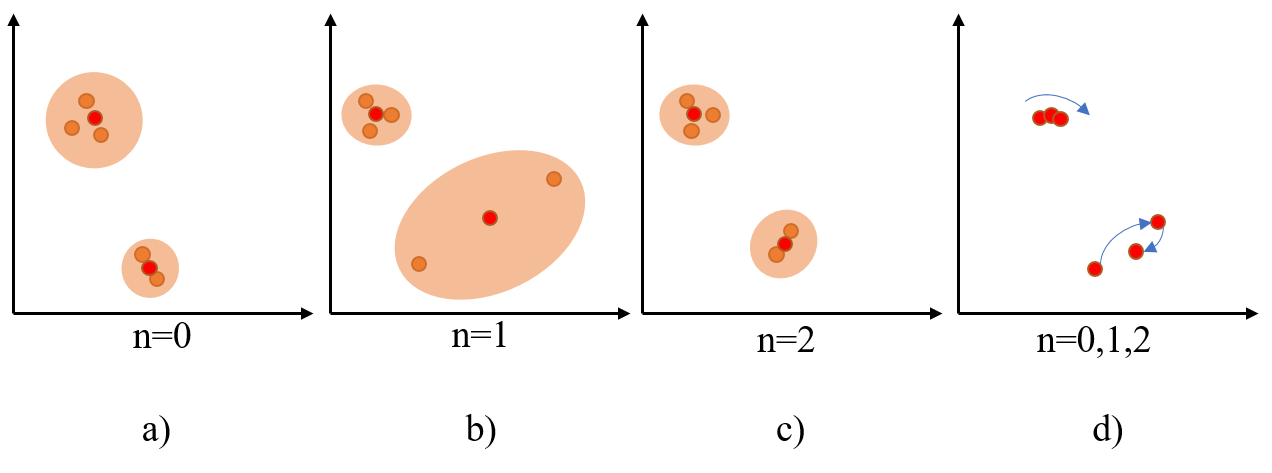}
\caption{Illustration of the failure case of the mid-hip clustering. From left to right: a), b), c) three continuous frames with the corresponding frame index $n=\left \{ 0, 1, 2 \right \}$, d) trajectories of mid-hips in three frames.}
\label{fig:jc_cluster_bad_case}
\end{figure}

\paragraph{Discussion} On possible failure case is depicted in Fig.~\ref{fig:jc_cluster_bad_case}. We illustrate it in a two-dimensional space. Similar to Fig.~\ref{fig:jc_cluster}, the orange dots represent the candidates and the red dots represent the cluster centroids. As shown in Fig.~\ref{fig:jc_cluster_bad_case}, there are three continuous frames in $n=\left \{ 0, 1, 2 \right \}$. Blue arrows illustrate two trajectories of the cluster centroids in the right sub-figure. We can find that, one trajectory is not smooth. Generally, this is because the performance of the KMeans algorithm tends to be affected by imbalanced data. In cases where mid-hip candidates are widely dispersed, the location of the individual centre can be unstable across frames. More advanced clustering algorithms can be employed to ameliorate the imbalanced data bias. Alongside this, object-tracking algorithms can be also employed to enhance the continuity of the Joint Cloud subset's centre across frames. In our case, the aforementioned scenario does not arise.

\setlength{\tabcolsep}{10pt}
\begin{table}[htbp]
\centering
\caption{Distance thresholds in metres for each joint type are listed here. We calculate the distance between the joint candidate and the individual centre.}
\label{table:jc_th}
\begin{tabular}{cc|cc}
\hline
\textbf{Joint type} & \textbf{Threshold} & \textbf{Joint type} & \textbf{Threshold} \\ \hline
Nose       & 0.85      & Neck       & 0.7       \\
RShoulder  & 0.7       & RElbow     & 0.8       \\
RWrist     & 0.8       & LShoulder  & 0.7       \\
LElbow     & 0.7       & LWrist     & 0.8       \\
MidHip     & 0.3       & RHip       & 0.3       \\
RKnee      & 0.6       & RAnkle     & 1.0       \\
LHip       & 0.3       & LKnee      & 0.6       \\
LAnkle     & 1.0       & REye       & 0.9       \\
LEye       & 0.9       & REar       & 0.85      \\
LEar       & 0.85      & LBigToe    & 1.0       \\
LSmallToe  & 1.0       & LHeel      & 1.0       \\
RBigToe    & 1.0       & RSmallToe  & 1.0       \\
RHeel      & 1.0       & -          & -         \\ \hline
\end{tabular}
\end{table}

\begin{figure}
\centering
\includegraphics[width=0.98\linewidth]{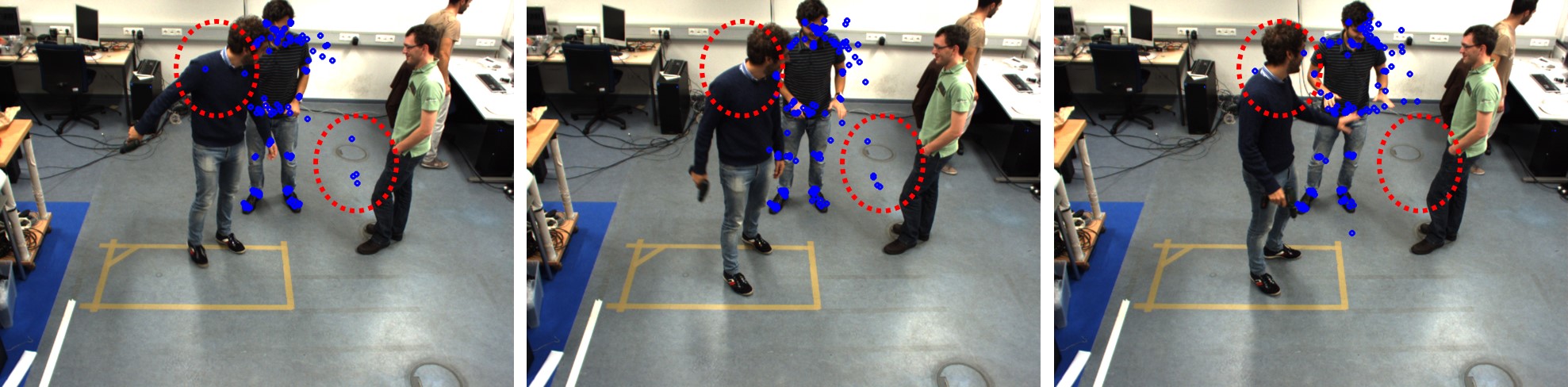}
\caption{Illustration of one individual's Joint Cloud in three continuous frames. Blue dots indicate the candidates that are projected back to 2D. Red dash lines highlight the outliers.}
\label{fig:jc_thr}
\end{figure}

\subsubsection{Thresholding for each individual}\label{sec:thres} 

For each cluster centroid (a.k.a. individual centre), we adopt several thresholds to filter outlier candidates and assign the remaining candidates to the current individual subset. Specifically, we first calculate the Euler distance between the candidate and the current individual centre. If the distance is bigger than the given threshold, we next filter this candidate. Table~\ref{table:jc_th} lists the detailed thresholds utilised in the main paper. 

\paragraph{Discussion} The above processes are intended to reduce the Joint Cloud's complexity. Nonetheless, one possible situation is that two individuals stand close and interact, rough thresholds may assign redundant candidates to each individual. This situation can be depicted as Fig~\ref{fig:jc_thr}. In Fig~\ref{fig:jc_thr}, there are three continuous Joint Cloud reprojections. While the outliers highlighted by red dash lines are not filtered correctly, we can still identify potential outliers through discontinuous patterns. To select the potentially valid candidates from the noisy and redundant Joint Cloud, we involve the spatial and temporal information: skeletal structure, and joint trajectory. The redundant candidates will lead to a false skeleton in every frame and discontinuous trajectories across the frames, and our proposed framework can distinguish them.


\subsection{Intuitive visual examples}

\begin{figure}
\centering
\begin{subfigure}[b]{0.46\textwidth}
\centering
\includegraphics[width=\textwidth]{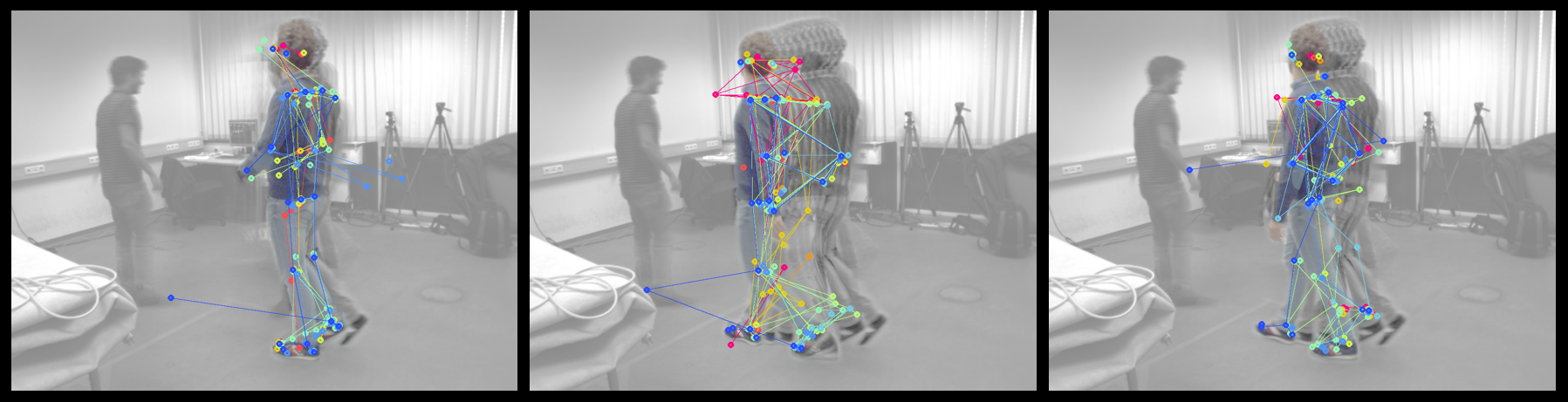}
\caption{Skeletal structures in all view pair combinations.}
\label{fig:jc_skel}
\end{subfigure}
\hfill
\begin{subfigure}[b]{0.46\textwidth}
\centering
\includegraphics[width=\textwidth]{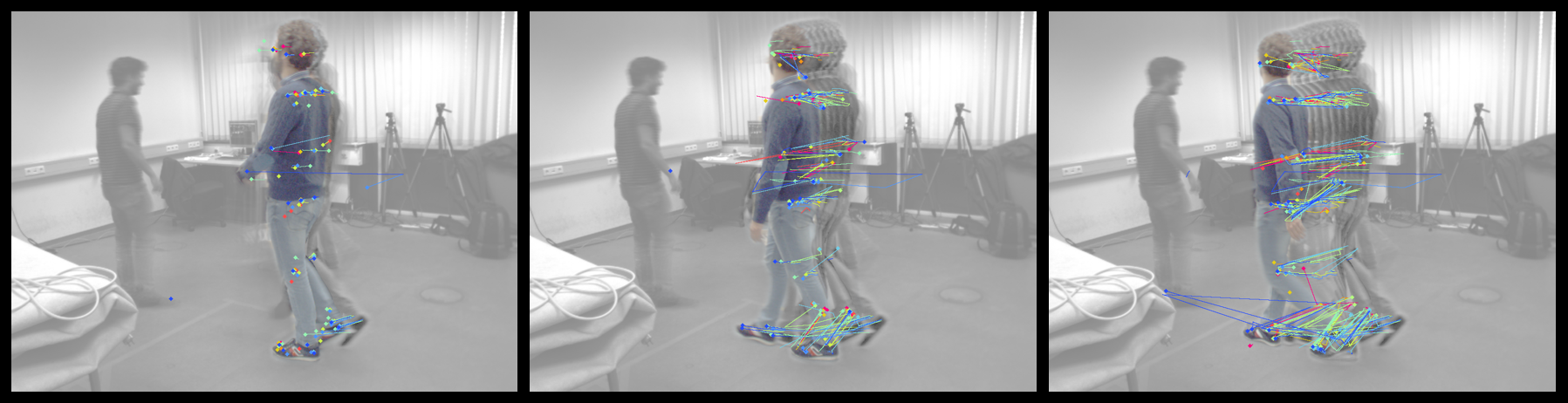}
\caption{Trajectories in all view pair combinations.}
\label{fig:jc_traj}
\end{subfigure}
\hfill
\caption{Demonstration of skeletal and trajectile information for one individual in Joint Cloud. Different colours represent different view pair combinations.}
\label{fig:jc_exmp}
\end{figure}

We provide intuitive visual examples to demonstrate the trajectories and skeletal structures in the proposed Joint Cloud. We want to justify that, in the Joint Cloud, we can select candidates to form a complete movement based on joint trajectories, skeletal structures and angular perspectives. Therefore a Transformer can also deeply explore the correlations among the trajectile, skeletal structural and angular information derived from Joint Cloud and regress those features to the 3D motion. In Fig.~\ref{fig:jc_exmp}, we present the skeletal and trajectile data for one individual in ten frames, with a variety of colour-coded view pairs to differentiate between them. In Fig.~\ref{fig:jc_skel}, we connect all possible limbs for every frame, thus revealing an approximate body skeletal structure. Similarly, in Fig.~\ref{fig:jc_traj}, we connect all joints in the same type to indicate their corresponding trajectories, thus revealing a rough body movement. Therefore, Joint Cloud can provide enough information to explore the implicit movement representations. With the help of the Transformer, we can regress those implicit representations to the final motion.

\begin{figure}
\centering
\includegraphics[width=0.98\linewidth]{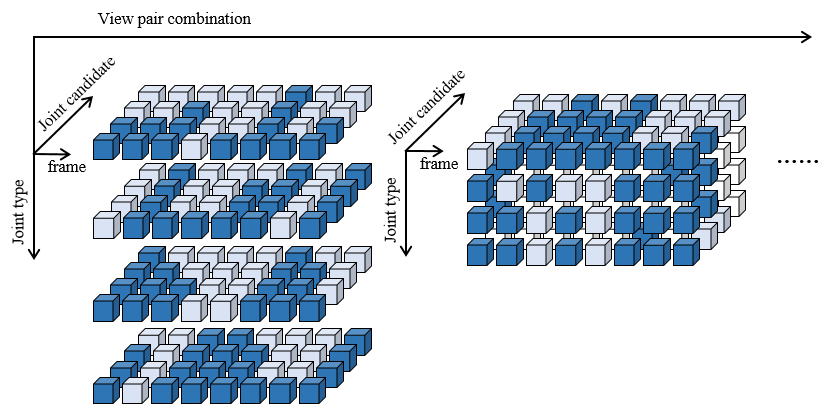}
\caption{Demonstration of the data arrangement of the Joint Cloud. Blue-coloured cubes and white-coloured cubes represent the candidates and zero padding respectively.}
\label{fig:jc_data}
\end{figure}

\subsection{Data arrangement}

As a supplementary explanation to the padding process described in Sec. 3.2 of the main paper, we visualize this process in Fig.~\ref{fig:jc_data}. Recall that the padded Joint Cloud is organized as a 5-dimensional matrix $\mathcal{J} \in \mathbb{R}^{N \times \dot{V} \times \dot{M} \times  I \times 3}$. If we ignore the last dimension, we can illustrate the matrix in Fig.~\ref{fig:jc_data}. Every blue-coloured cube represents one joint candidate, and every white-coloured cube represents zero padding. In practice, the maximum limit for the joint candidate count is restricted to four.

\section{Network structure}

As introduced in the main paper, the proposed Joint Cloud Selection and Aggregation Transformer (JCSAT) consists of three Transformer-based encoders for trajectile, skeletal structural and angular cross-embedding. 
Each encoder is a vanilla Visual Transformer (ViT)~\cite{dosovitskiy2020image} consisting of multiheaded self-attention blocks (MSAs) and feed-forward network blocks (FFNs). Layernorm (LN) is applied before MSAs and FFNs. Residual connections are employed after every block.
JCSAT takes Joint Cloud as input and treats every joint candidate as an independent token encoded by the Trajectory Encoder and Structure Encoder in parallel. After encoding, we propose an Optimal Token Attention Path (OTAP) module that differentiates and optimally selects reliable and representative candidate token features from the Trajectory Encoder and Structure Encoder respectively. Next, we sum up trajectile features and skeletal structural features on the view dimension (in the main paper, we also refer to features as representations). So for each view pair combination, there is one motion feature to represent the body motion. Then, we encode the motion features by the View Encoder followed by an OTAP module to select one view's motion feature among all view pair combinations. Last, an MLP prediction head is used to regress the motion feature to 3D movements.
Table~\ref{table:jcsat} and Table~\ref{table:otap} show the details of the JCSAT structure. We will release the whole code of our proposed framework as well as the newly collected multi-view multi-person labelled motion capture dataset after publication.

\setlength{\tabcolsep}{15pt}
\begin{table}[htbp]
\centering
\caption{The detailed structure of the proposed Joint Cloud Selection and Aggregation Transformer (JCSAT) network. `Num' denotes the number of layers. `Dim' denotes the corresponding dimension.}
\label{table:jcsat}
\begin{tabular}{lc}
\hline
\textbf{Trajectory Encoder} & Num \\ \hline
In Dim & 3 \\
Enc Dim & 256 \\
Enc Depth & 16 \\
MSA Head & 16 \\
MSA Dim per Head & 16 \\
FFN Dim & 256 \\ \hline
\textbf{Structure Encoder} & Num \\ \hline
In Dim & 3 \\
Enc Dim & 256 \\
Enc Depth & 16 \\
MSA Head & 16 \\
MSA Dim per Head & 16 \\
FFN Dim & 256 \\ \hline
\textbf{View Encoder} & Num \\ \hline
In Dim & 256 \\
Enc Dim & 32 \\
Enc Depth & 8 \\
MSA Head & 8 \\
MSA Dim per Head & 4 \\
FFN Dim & 3 \\ \hline
\end{tabular}
\end{table}

\setlength{\tabcolsep}{15pt}
\begin{table}[htbp]
\centering
\caption{The detailed structure of the proposed Optimal Token Aggregation Path (OTAP) module. We implement it by a trainable optimal transport technique~\cite{mialon2020trainable}. `Num' denotes the number of layers. `Dim' denotes the corresponding dimension.}
\label{table:otap}
\begin{tabular}{lc}
\hline
\textbf{T-OTAP} & Num \\ \hline
In Dim & 256 \\
MSA Head & 1 \\
Log Domain & True \\
Out Dim & 25 \\ \hline
\textbf{S-OTAP} & Num \\ \hline
In Dim & 256 \\
MSA Head & 1 \\
Log Domain & True \\
Out Dim & 10 \\ \hline
\textbf{V-OTAP} & Num \\ \hline
In Dim & 32 \\
MSA Head & 1 \\
Log Domain & True \\
Out Dim & 1 \\ \hline
\end{tabular}
\end{table}

\subsection{Joint type conversion}

As discussed in Sec. 3.2.2 of the main paper, we utilize an additional MLP layer after the prediction head to convert the predicted movement into the ground-truth joint type. Inside, there are two linear layers that convert the input dimension from 25 to 20 and 20 to 15 sequentially.

\section{Training details}

To easily reproduce our result, we provide the training specifications in Table \ref{table:training_config}. Moreover, following \cite{dosovitskiy2020image,he2022masked,jiang2022dual}, we adopt the linear learning rate scaling rule~\cite{goyal2017accurate}, i.e., $\mathit{lr}=\mathit{base\:  lr} \times \mathit{batch \: size} / 256$.

\begin{table}[htbp]
\centering
\caption{Specifications for network training.}
\label{table:training_config}
\begin{tabular}{lc}
\hline
\textbf{Specifications}                                 &                                 \\ \hline
optimizer                                               & AdamW~\cite{loshchilov2017decoupled}   \\
optimizer momentum                                      & $\beta_1, \beta_2=0.9, 0.999$   \\
weight decay                                            & 0.05                            \\
learning rate schedule                                  & cosine decay~\cite{loshchilov2016sgdr} \\
warmup epochs~\cite{goyal2017accurate} & 30                              \\
gradient clipping                                       & 0.02                            \\
drop out                                                & No     \\
base learning rate                                      & 2e-4                            \\
batch size                                              & 5                               \\
epoch size                                              & 120                            \\
\hline
\end{tabular}
\end{table}

\section{Additional results}

We offer supplementary visual comparisons with Zhang et al.~\cite{zhang20204d} (CVPR'20), Dong et al.~\cite{dong2021fast} (TPAMI'21) and Jiang et al.~\cite{jiang2022dual} (MM'22) on the Shelf dataset and our newly collected multi-view multi-person motion capture dataset. We highlight the false reconstruction with red dotted boxes. Same to the main paper, the visualization of Joint Cloud for one individual is included for every frame sample. Furthermore, we provide synthetic videos for continuous evaluation. 

\begin{figure*}[t]
\centering
\includegraphics[width=0.85\linewidth]{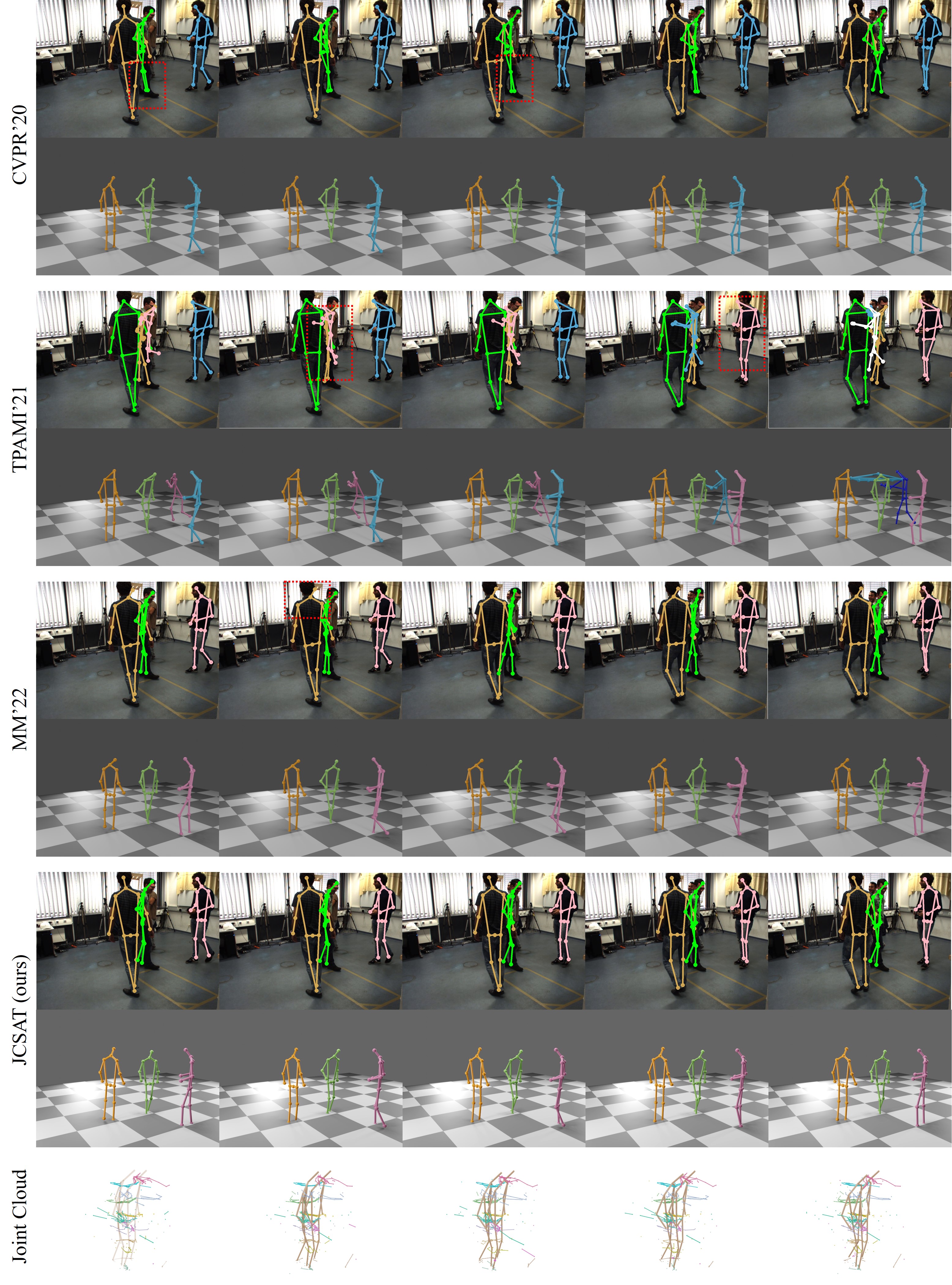}
\caption{Visual comparisons on the Shelf dataset in five continuous frames (214-218). We demonstrate 3D poses and the corresponding reprojection on the 5th view. From top to bottom, Zhang et al.~\cite{zhang20204d} (CVPR'20), Dong et al.~\cite{dong2021fast} (TPAMI'21), Jiang et al.~\cite{jiang2022dual} (MM'22), our proposed framework (JCSAT) and Joint Cloud. We highlight the false reconstruction with red dotted boxes.}
\label{fig:shelf_214_218}
\end{figure*}
\begin{figure*}[t]
\centering
\includegraphics[width=0.85\linewidth]{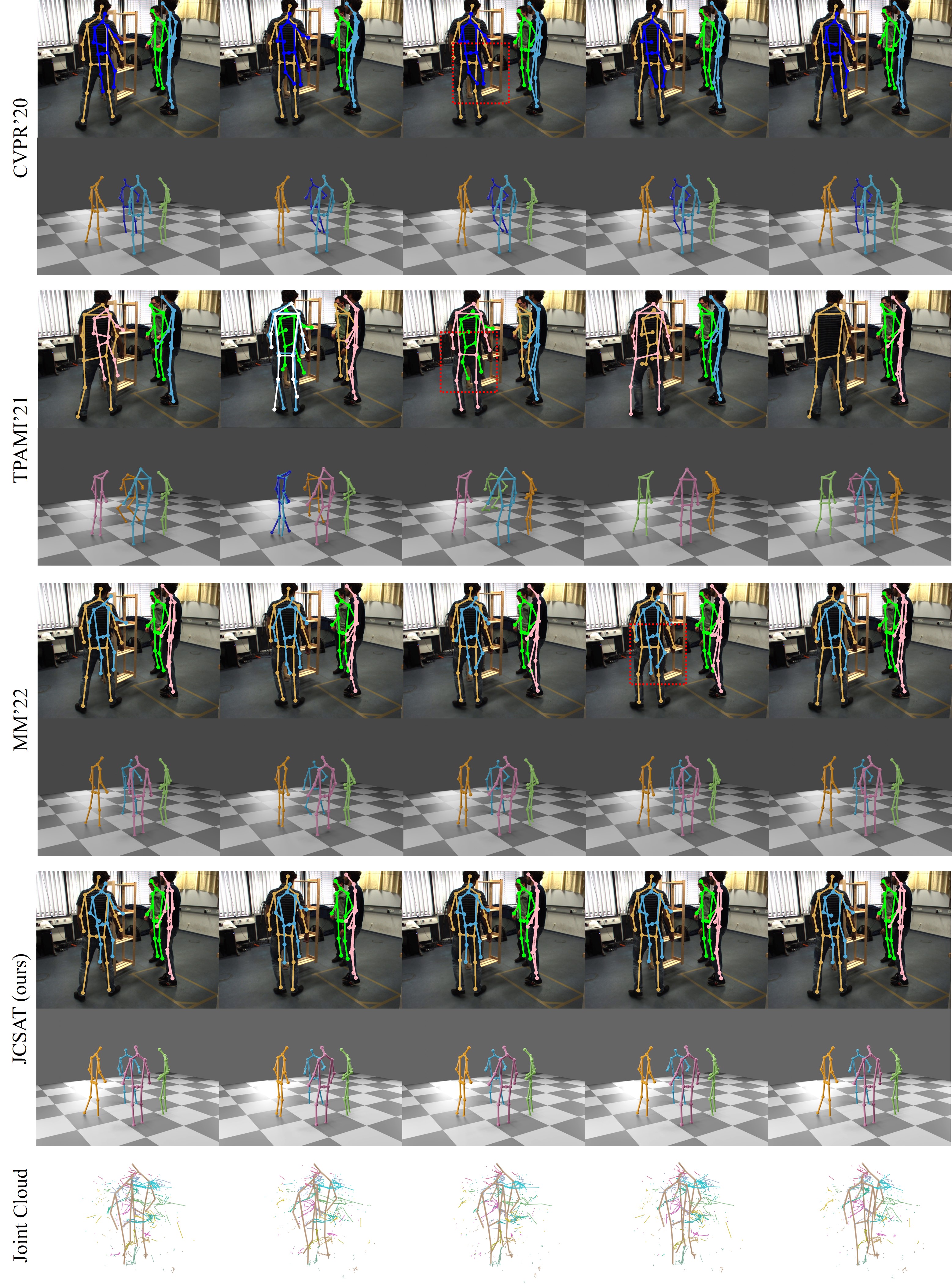}
\caption{Visual comparisons on the Shelf dataset in five continuous frames (246-250). We demonstrate 3D poses and the corresponding reprojection on the 5th view. From top to bottom, Zhang et al.~\cite{zhang20204d} (CVPR'20), Dong et al.~\cite{dong2021fast} (TPAMI'21), Jiang et al.~\cite{jiang2022dual} (MM'22), our proposed framework (JCSAT) and Joint Cloud. We highlight the false reconstruction with red dotted boxes.}
\label{fig:shelf_246_250}
\end{figure*}

Fig.~\ref{fig:shelf_214_218} and Fig.~\ref{fig:shelf_246_250} demonstrate the results of two continuous clips on the Shelf dataset. In Fig.~\ref{fig:shelf_214_218}, let's focus on the second and third individuals which are under heavy conclusion. \cite{zhang20204d} reconstructs most of the correct body parts while the left foot is wrongly produced. \cite{dong2021fast} reconstructs inconsistent identifications (the third individual's colour changes through frames). \cite{jiang2022dual} predicts the imprecise head location. Our framework produces more accurate results in terms of identification consistency (don't change through frames) and joint location precision (the second individual's foot location is relatively accurate). In Fig.~\ref{fig:shelf_246_250}, \cite{zhang20204d,dong2021fast,jiang2022dual} struggle in reconstructing the third individual's movement. Since the third individual is under heavy occlusion, the locations of the feet are badly reconstructed. Also \cite{dong2021fast} suffers from inconsistent individual identification through frames (the skeleton colour is changed). But in our framework, the proposed JCSAT explores the joint trajectory and skeletal structure information. The occluded joint part can be regressed from the former and latter frames so that our framework outperforms others.

\begin{figure*}[t]
\centering
\includegraphics[width=0.98\linewidth]{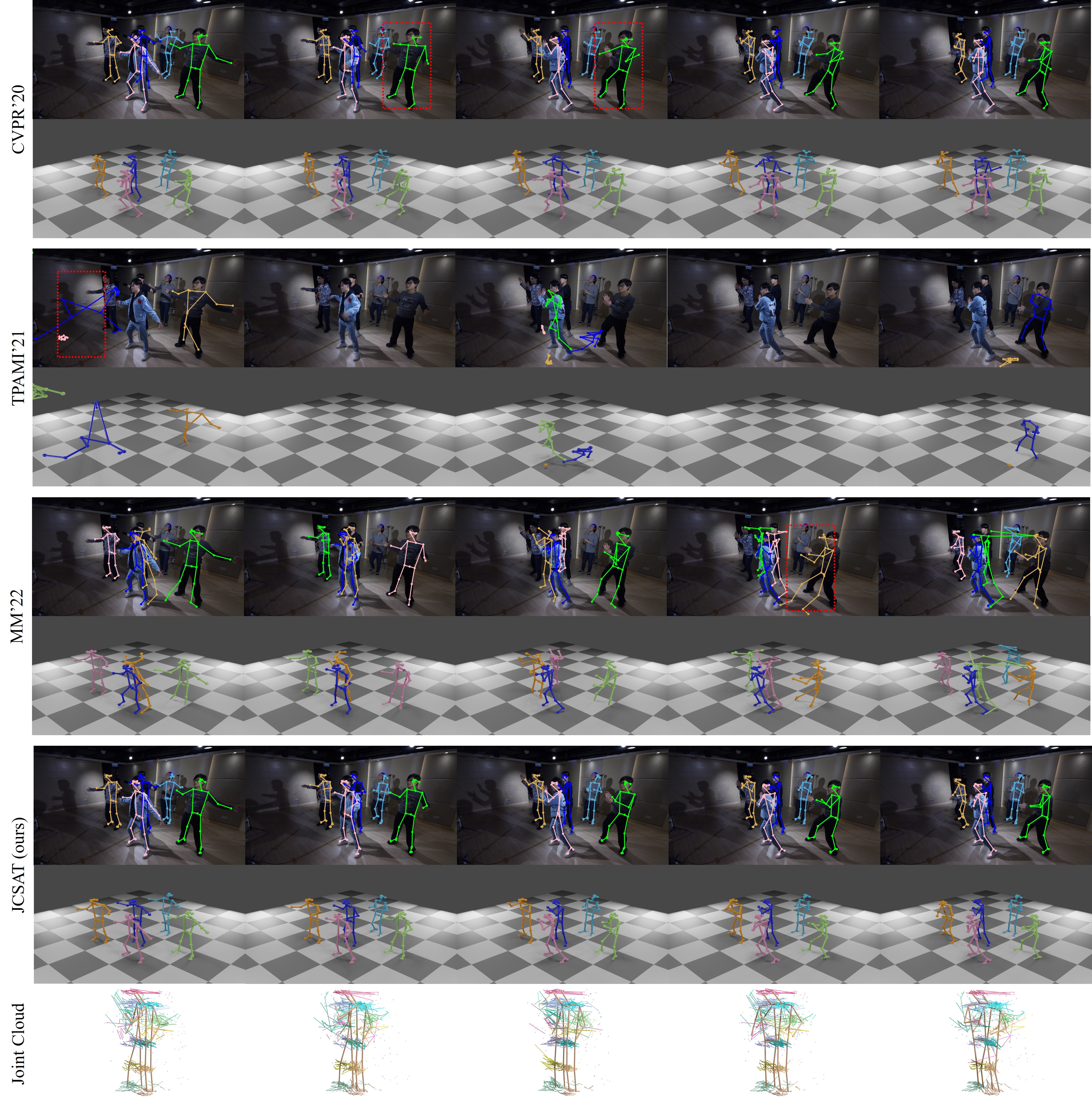}
\caption{Visual comparisons on our newly collected dataset in five continuous frames (30-34). We demonstrate 3D poses and the corresponding reprojection on the 5th view. From top to bottom, Zhang et al.~\cite{zhang20204d} (CVPR'20), Dong et al.~\cite{dong2021fast} (TPAMI'21), Jiang et al.~\cite{jiang2022dual} (MM'22), our proposed framework (JCSAT) and Joint Cloud. We highlight the false reconstruction with red dotted boxes.}
\label{fig:coop8_30_34}
\end{figure*}
\begin{figure*}[t]
\centering
\includegraphics[width=0.98\linewidth]{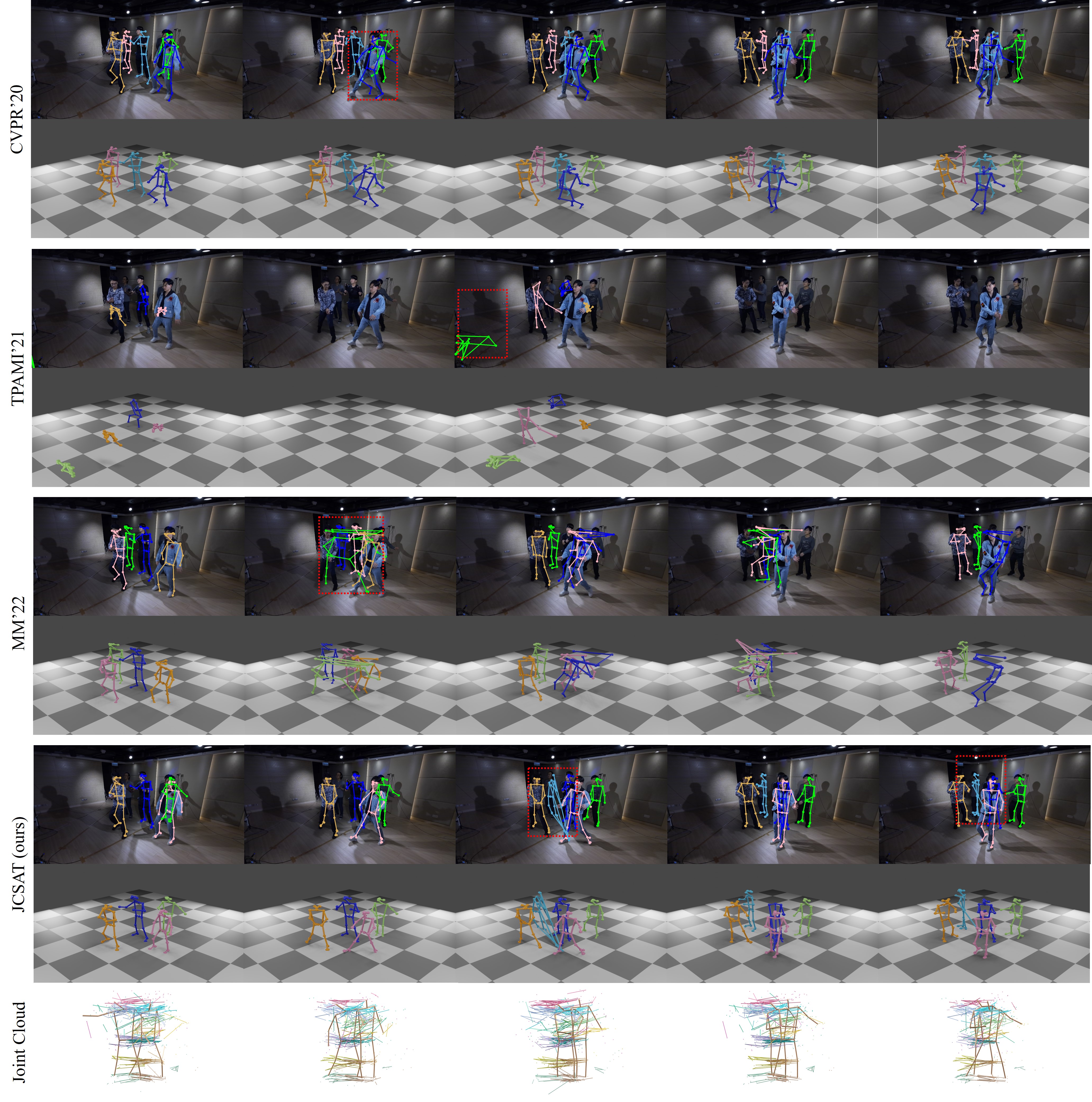}
\caption{Visual comparisons on our newly collected dataset in five continuous frames (235-239). We demonstrate 3D poses and the corresponding reprojection on the 5th view. From top to bottom, Zhang et al.~\cite{zhang20204d} (CVPR'20), Dong et al.~\cite{dong2021fast} (TPAMI'21), Jiang et al.~\cite{jiang2022dual} (MM'22), our proposed framework (JCSAT) and Joint Cloud. We highlight the false reconstruction with red dotted boxes.}
\label{fig:coop8_235_239}
\end{figure*}

Fig.~\ref{fig:coop8_30_34} and Fig.~\ref{fig:coop8_235_239} demonstrate the results of two continuous clips on our newly collected dataset. This dataset records five individuals' dancing movements including turns and rotations with heavy occlusions and ambiguities leading to a challenging scenario. Thus we want to validate our proposed framework that JCSAT learns to regress the motion via trajectories and skeletal structures and also show its superiority among state-of-the-art frameworks. In Fig.~\ref{fig:coop8_30_34}, \cite{zhang20204d} reconstructs discontinuous movements while the tracking energy function gives all individuals consistent identities. \cite{jiang2022dual} fail to reconstruct correct movements due to its simple re-id algorithm. The re-id algorithm is unable to differentiate between two individuals with similar appearances. Our framework regresses more accurate movements. \cite{dong2021fast} fails to reconstruct most of the movements. We believe the failure reason is similar to \cite{jiang2022dual}'s. That is \cite{dong2021fast} wrongly re-identify each detection across views. Besides, we find that \cite{dong2021fast}'s 2D pose detector is prone to detect humanoid shadows as poses. In other words, \cite{dong2021fast} is sensitive to the appearance of detections. If the detections share similar appearances (e.g. dark-coloured outlooking in a weak light environment and the grey shadows), the re-id algorithm will fail.
Despite this, in Fig.~\ref{fig:coop8_235_239}, our framework fails in some frames in terms of the light-blue-coloured individual. This is because, in the first, JCSAT doesn't detect this individual. The trajectile information is missing. \cite{zhang20204d} reconstructs incorrect arms for the dark-blue-coloured individual. Overall, ours performs better than \cite{dong2021fast,zhang20204d,jiang2022dual} in this scene.

\end{document}